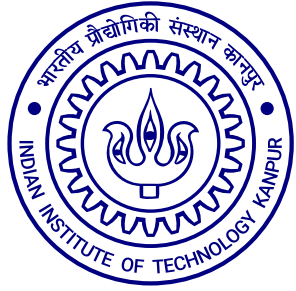

# GEOMETRIC UNDERSTANDING OF SKETCHES

A Thesis Submitted

In Partial Fulfilment of the Requirements

for the Degree of

**DOCTOR OF PHILOSOPHY**

by

Raghav Brahmadesam Venkataramaiyer

Roll No. 14119266

to the

Department of Design

**INDIAN INSTITUTE OF TECHNOLOGY KANPUR**

March 15, 2022

# CERTIFICATE

It is certified that the work contained in the thesis titled "**Geometric Understanding of Sketches**," by "*Raghav Brahmadesam Venkataramaiyer*" has been carried out under my/our supervision and that this work has not been submitted elsewhere for a degree.

March 15, 2022          Dr. Vinay P. Namboodiri          Dr. Satyaki Roy

CSE IITK                          DP IITK

(Supervisor)                          (Administrative Supervisor)

# DECLARATION

This is to certify that the thesis titled "**Geometric Understanding of Sketches**," has been authored by me. It presents the research conducted by me under the supervision of *Dr. Vinay P. Namboodiri.*

To the best of my knowledge, it is an original work both in terms of research content and narrative, and has not been submitted elsewhere, in part or in full, for a degree. Further, due credit has been attributed to the relevant state-of-the-art and collaborations (if any) with appropriate citations and acknowledgements, in line with established norms and practices.

March 15, 2022

Raghav Brahmadesam Venkataramaiyer
Doctor of Philosophy

Department of Design
Indian Institute of Technology Kanpur
Kanpur 208016

# ABSTRACT/SYNOPSIS

*Name of Student:* Raghav Brahmadesam Venkataramaiyer     *Roll no:* 14119266

*Degree:* Doctor of Philosophy     *Department:* Department of Design
*Thesis Title:* Geometric Understanding of Sketches

*Name(s) of Thesis Supervisors:*
   1. Dr. Vinay P. Namboodiri

Sketching is used as a ubiquitous tool of expression by novices and experts alike. In this thesis I explore two methods that help a system provide a geometric machine-understanding of sketches, and in-turn help a user accomplish a downstream task.

The first work deals with interpretation of a 2D-line drawing as a graph structure, and also illustrates its effectiveness through its physical reconstruction by a robot. We setup a two-step pipeline to solve the problem. Formerly, we estimate the vertices of the graph with sub-pixel level accuracy. We achieve this using a combination of deep convolutional neural networks learned under a supervised setting for pixel-level estimation followed by the connected component analysis for clustering. Later we follow it up with a feedback-loop-based edge estimation method. To complement the graph-interpretation, we further perform data-interchange to a robot legible ASCII format, and thus teach a robot to replicate a line drawing.

In the second work, we test the 3D-geometric understanding of a sketch-based system without explicit access to the information about 3D-geometry. The objective is to complete a contour-like sketch of a 3D-object, with illumination and texture information. We propose a data-driven approach to learn a conditional distribution modelled as deep convolutional neural networks to be trained under an adversarial setting; and we validate it against a human-in-the-loop. The method itself is further supported by synthetic data generation using constructive solid geometry following a standard graphics pipeline. In order to validate the efficacy of our method, we design a user-interface plugged into a popular sketch-based workflow, and setup a simple task-based exercise, for an artist. Thereafter, we also discover that form-exploration is an additional utility of our application.

Using the two methods, validated against a real-world setup, we set a precedent for an arguably robust machine understanding of geometric forms expressed as sketches, in the former for 2D geometry in the context of a robot-based task, and in the latter for 3D geometry in the live context of a sketch-completion task.

इदम् पितृभ्याम्। इदम् गुरुभ्यो। इदम् देवाय। इदन्नमम॥

x

# Acknowledgement

I bow down before all my teachers starting with *Mrs. Rekha Sharma, Mrs. Manjeet Sial* from school; *Er. Aditya Kumar, Er. Neeraj Kumar* from preparation days; the professors of *IIT Roorkee*, and the professors of *IIT Kanpur*. Thank you so much... for you shaped me.

My gratitude is due, to all the resources, books, blogs, online courses, libraries and catalogues, that have helped me transition from the field of architecture to AI. To *Brian Strang, Cormen-Leiserson-Rivest-Stein, Erik Demaine, Srini Devadas, Dimitri Bertsekas, David McKay, Tim Roughgarden, Daphne Koller, Grant Sanderson, Kernighan & Ritchie, Bjarne Stroustrup, Herb Sutter* — Thank you all for making this journey possible for me.

A lot of softwares have helped me express myself at different fora, or even to take notes. *Knuth/Lamport* for LATEX, *Stallman* for GNU and Emacs, *Torvalds* for Linux and Git, *Dominic Carsten* for Org Mode, *John Kitchin* for Org Ref, *Jethro Kuan* for Org Roam, *Bram Moolenaar* for VIm, *Guido von Rossum* for Python, *Ton Roosendaal* for Blender — I express my gratitude to you all and your collaborators, and also to the authors, developers and maintainers of Archlinux, I3WM, Magit, GIMP, Inkscape, Google Scholar Tools, PyTorch Framework, PyTorch Lightning — for I own my workflows because of you.

This thesis is a result of collaboration with great minds, *Subham Kumar, Abhishek Joshi* and *Saisha Narang*; at the Deltalab: *Sarvesh, Badri, Kurmi, Puma, Munnu, Ravindra, Samik, Avideep, Soumya*; at the Helicopter: *Aravind, Ganesh, Priyanka, Mj, Sreerag, Amarish, Bongo, Abhishek, Sarvesh* and *Anubhav*. Thank you all for being with me.

I may count all the beads, but it won't be a necklace without the thread, that is you, my advisors, *Dr. Vinay Namboodiri* and *Dr. Satyaki Roy*. You have seen me float through the better and the worse. Thank you.

I am fortunate to have my family — *Amma-Appa-Apoo-Kaku-Preetu* — *my friends* and *relatives.* Thank you all so much... for you define me.

# Contents

# List of Figures

# List of Tables

# Introduction

Geometry with its strong mathematical ground, provides far reaching consequences for many a field like, and does so for almost as early as humanity, or even earlier. Geometric patterns have been symbolic of human civilizations, from as early as the third millennium BC across the globe in the Stonehenge, the Harappan cities, the pyramids of Egypt and countless others. Not only humans, we also see geometric expressions amongst the plant and animal kingdom, for example the geometry of cauliflower, the flock of starlings, and erect position of squirrels on a predation alarm call. We also see, not surprisingly, the geometric constructs in the essence of more ephemeral human expressions of sketches, whose intuitive expressions have driven research in the psycho-social domain for a long time. More recently, we see scientists interested in the recognition and parsing of sketches from a computational point of view.

One dimensional geometry is trivial and overlaps with arithmetic operations on the number line. The impact of geometry is, however, significantly observable with the second and third dimension. Here I would like to highlight the fact that sketches drawn as a raster graphic have far reaching consequences when interpreted as geometric representations.

For example we have seen Felzenszwalb's work on polygonal shape estimation of planar regions, using an efficient elimination method [Fel03]. And we have also seen a series of works in the domain of interactivity by Igarashi, towards creative exploration with rotund objects observable to humans as sketches, and processed by the machine as geometry of the third dimension [IMT99; MI07]. In general, geometric constructs tend to compose a geometry using "simple" atomic elements, for example a closed planar curve may be approximated as a CSG tree with simple atomic elements of circles and polygons as its leaf nodes [Män87]. This has the advantages of the standard create-read-update-delete (CRUD) operations over the geometry. So editing, transforming, repeating, often with symmetrical constructs provide efficient way to explore higher order visual forms.

However, study suggest that a typical creative user struggles to accommodate this computationally efficient structure into her natural workflow and often finds it uncomfortable. Perhaps, maintaining a structured mental model of complex geometry along with a creative exploration is paradoxical and counter-intuitive — but that's an open question.

In contrast to the geometric modelling paradigm, what it takes to sketch, is just a planar medium that can afford leaving an impression when in hard contact with a pointed object, for example on a beach, kids enjoying their vacation leave impressions on sand using a stick. Or someone doodling on the back of a paper napkin while in a coffee shop. Google's experiment that resulted in the *Quick-draw* dataset [HE17] encouraged users to draw with fingers on a mobile phone. More sophisticated businessmen would use a pen tablet to jot down their ideas. It won't be far-fetched to imagine using the advances in augmented and virtual reality to push the boundaries of sketching as an activity in future, say sketching in thin air! Arguably, sketching is a low-cost affair, and so is recording a sketch!

Sketches are omnipresent. They span across industries, across the design life-cycle, across history. We look at the sleek curve on the bonnet of a car and we know someone had sketched it to perfection. We see the Guggenheim museum at Bilbao, and know that Zaha Hadid had communicated her idea to the digital artists as a sketch. When we see the floor of a toilet sloping down towards the trap, we know that someone had detailed it out in a sketch/ drawing. We look at the detail of the root canal while waiting in a dentist's lobby, and we know that there was someone who diligently had observed and drawn the microscopic details of a tooth. We look inquisitively for our station on the map, while travelling in a rapid transit, time and again, and someday we would wonder that someone should have envisaged it as a sketch. In the design lifecycle of a product too, sketches are used to communicate, explore, argue and validate the ideas during the design phase; to document the user responses in testing phase; and to document the product itself after completion vis-à-vis planning and emergencies a posteriori.

Though we may see an ease of use for a sketch from the point of orientation towards sketching as a method in a workflow, geometry too has its advantages. Geometry is a rigorous science; it is mathematical in nature with formalism guiding the progress, *e.g.* a whole field of euclidean geometry, circular and spherical trigonometry rises from a simple observation that a right triangle may be understood as a composition of a base, a perpendicular and a hypotenuse with respective lengths $b$, $p$ and $h$ and related mutually as $b^2 + p^2 = h^2$. Geometrical constructs, thus, provide for a great tool for rigorous analysis. However, the rigorous process of geometrical construction is arguably an intensive cognitive process often in conflict with exploratory processes like problem solving.

Sketching on the other hand is visuo-spatial activity, which is guided by the artists mental model of physical world. In contrast to the formalism and rigour of geometrical constructs, sketching provides freedom for unconstrained exploration. As an old proverb goes, "a picture is worth a thousand words." For example, in case of forensics, sketching skills of one person gives shape to someone else's imagination. The same argument holds good for technical sketches drawings, *e.g.* the machine assembly drawings which incorporate the temporal dimension also in the same sketch.

History suggests that sketching has been used as a medium of communication from as early as undeciphered history, in the petroglyphs of Bhimbetka (see Fig. 1.1), when early men lived in rock shelters, similar in nature to the aboriginal rock art of Australia and paintings in the prehistoric caves of Lascaux. Common themes include animals, dancing and hunting. These paintings help us to establish a communication channel across time; a very long time, perhaps.

But sketches and geometry go hand in hand, as in the Renaissance humanism, that irrespective of the opinion of the church, promoted a philosophy of science, experiential observation, geometry, and mathematics. Take for example, a fifteenth century Roman painter who, besides painting, was also known for his notes on anatomy, astronomy, botany and cartography. His name, Leonardo da Vinci, evokes respect for his epitomized ideal ever since.

In the turn of the eighteenth century, Newton's laws of motion revolutionised the philosophical pursuit. We were able to understand the planetary motion, tides, comets, equinoxes and to assert heliocentricity using the same principles as those in the motion of objects on Earth! It is not unimaginable that before arriving at the *Principia*, Newton used sketches to observe, document and analyse the physical phenomena, when evidently, we grow up learning and practising

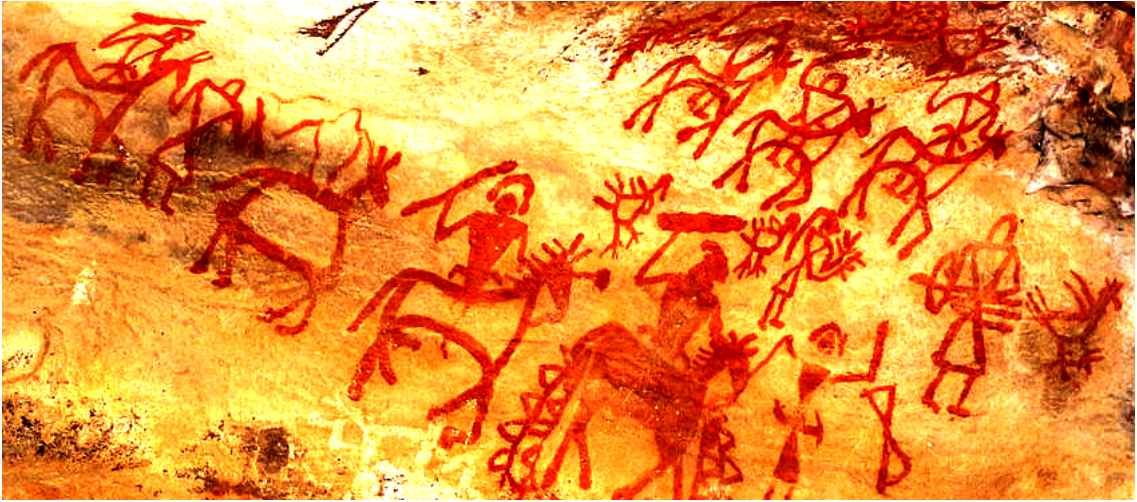

Figure 1.1: Prehistoric line art expressions discovered in the caves of Bhimbetka. Sketching is a human preoccupation and has invariably been used a medium to convey ideas and information across time and space.

the same principles using free-body diagrams.

Whether as novices, or as experts at sketching, the students of science, use the free-body diagrams with the essence and purpose of analysis across scale and context as a tool. The subject of study may be a box on a slope or a swinging pendulum or the earth revolving around the sun. The objective may be to understand the projectile of a cricket ball, or the collision of snooker balls or the rolling of a cycle wheel. Regardless, we use the same visual language as an abstraction of the real world; a language that is simple enough to capture all the relevant details required to analyse, define and solve the problem — *not any simpler*; a language that filters out all the noise pertinent to the real world, from the point of view of problem solving.

The use of sketch as a language of documentation is undeniably useful in the sphere of design. A design exercise starts with research to arrive at a set of guiding principiles/ directives/ objectives. The further process may be seen as a feedback loop with ideation, exploration and validation. In the feed-forward phase, the designer ideates with a set of intangible inspirations and explores to create many a tangible but semi-concrete unfinished design alternatives, as a feed-forward. These alternatives are then evaluated and validated against the objectives set forth at the start of the exercise. The insights from the evaluation and validation act as a feedback and form the inspiration for the next iteration of the loop. With human-centric principles and objectives, the design loop becomes human-centric in nature. A sketch is the essential language of documentation and communication in this exercise.

The power of visual communication has, since, been recognised, and has given way to dedicated disciplines in the modern world, like engineering drawing, architectural graphics, technical drawing, classical animation, commercial illustration and so forth. Under such a specialisation, a student is equipped with specific-to-purpose geometric tools like a drafter, parallel-bar, set-squares, french curves etc. and is trained to achieve an acceptable precision. The acquired skill is an expertise in a specific visual language that is, perhaps, a fundamental building block to a domain-specific problem-solving and communication. A person knowing such a language may, on one hand, use sketching to express and analyse a problem and explore the solutions. While on the other hand, she may communicate the solution in a geometrically precise manner using the specialised tools. Modern tools for manual drafting were used to author documents in a domain-specific language, for example electrical drawing, structural drawing, manufacturing drawing, architectural drawing, good-for-construction drawing etc; thereby providing for multi-level precision, and allowing for correction/ updates.

The tools of modern era had given way to machines of today, where advances in computation provide lightning-fast feedback and sub-atomic precision. Advances in computational geometry and progress in commercial hardware led to widespread use of computer-aided design (CAD) tools. The significance of this shift may be observed in the advances documented in the respective industries of integrated circuits, manufacturing, construction and also in the communities of architecture, graphics and design. The efficient editing capabilities of a machine, combined with the lightning fast feedback using an electronic display, immensely enhances the iterative design process, and potentially registers a hyperbolic improvement in productivity when coupled with advances in the information technology.

With the introduction of geometric tools earlier, and later the computation-based improvisations of the same, there has been a leap in areas of geometry processing and graphical processing. However, the workflows till date rely on clean data input from the user. This limitation has been a cause of concern for the creative minds.

During my experience as a practising architect, I was responsible for providing drawings as per the users' requirements. A user provides his requirements to the architect, the functional part of which may be translated as room sizes and room relationships; but there also exist another significant part of style and personal preferences — for example fluidity, privacy and so forth — that remain a prerogative of the designer to implement and judge. While the former, may be thought of as an algorithmic exercise, and has been rightly attempted so, the latter has been an arena of human judgement; perhaps, thanks to the creative faculties of the visual thinker that helps to express one's mental model of the problem through sketches, and thus explore the solution space.

A description of this sketching activity may yield some insights. Sketching is rough in nature; meaning that it is a low precision projection of one's mental model. But it captures the essence; for example a sketch of a face captures the defining characteristics in a handful of strokes! And also that a sketch is likely to be *sprinkled* with a lot of low-intensity hazy lines, often critical to the user's journey to the final set of strokes. In my personal experience, my sketches often remind me of the inductive, analogical or counterfactual reasoning activity in hindsight, but are often hidden to the untrained eye. They are layered into different intensities and line-thicknesses, often fading into the background as noise.

A visual thinker is not limited by sketching in the traditional sense as the only mode of expression. It may be argued that CAD-based workflows provide an equally insightful visual feedback to aid exploration. But their meagre acceptance has been studied by Athavankar, and my observations finds resonance in his research on "Thinking Design" [Uda90], where design and its associated thought process have been analysed from a manipulo-visuo-spatial point of view. I quote an excerpt here as follows,

> "...It challenges the very idea of using standard primitives and Boolean operations to generate shapes! Any strategy that separates the feel of actual geometry and spatial movement from the process of shape generation and manipulation will not be complementing the thinking process. It is no wonder that the designers are not comfortable in using coordinates as input to generate shapes..."

"Using coordinates as input to generate shapes" is counter-intuitive. It may ordinarily seem that, to start with a clean slate, the first step is to draw a familiar shape, irrespective of the medium of expression. And for most of such "context-free" drawing exercises, the distinction between sketching and CAD-drawing, may dull out. But when we look at a contextual scenario, the difference should be appreciable. For example, in an existing theatre complex, if I intend to carve some space out for a support crew, I would start out with exploring options; say close to the green rooms, close to the service core, and so forth. In such an attempt a mark is made by a visual thinker on these locations. In case of a sketch, these marks are often subdued, like the layers of different intensities and line-thicknesses, which often fade into the background as noise. The visual form of these marks are a subconscious decision. But in case of a geometric construct, a definitive geometry, albeit as simple as a straight line, is a prerequisite to leave a mark. The difference lies where a decision of visual form has to be made. In case of a sketch, the decision is subconscious, whereas, in case of a geometric construct, a conscious decision has to be made.

Another difference lies in the nature of sketch refinement. Whereas a sketch is refined using overdraft, that is by drawing many a stroke over and over again, increasing the intensity of the impression on paper/ medium; and an eraser is sparingly used. It implies that the prior-to-refinement art is present like breadcrumb trail on the sketch itself. In contrast, the same set of operations in a geometric construction based workflow would create multiple geometry leading to redundancy. Hence the geometric workflows would rely on update and delete operations as much as on the create operations, depriving the user of any fading breadcrumbs in his art unlike a sketch. Also maintaining a mental model of the hierarchical construct of a geometry, with an increasingly complex structure is counter-intuitive, paradoxical, and a cognitively intense exercise.

Thus we see preference for sketch-based workflows and artefacts among the design circle during the exploration phase. However, The benefits of using CAD-based workflows far outweigh the weaknesses, and so, we see a prevalence of CAD models and drawings in other phases of the design life cycle. This leaves us with a gap of understanding the geometry from sketches and feeding the precise input to the CAD-based interfaces, which is currently filled-in by a design-aware skilled workforce. But the process is time consuming. This leads us to a gap where if the journey from sketch to a geometry is automated, we hope to have an end to end computational workflow, that may assist the design decisions with an improved feedback from sketches. In this thesis, we attempt to create **a mathematical model that takes a sketch made by a user as input and satisfactorily understands the underlying geometry.**

Traditionally, we observe the human machine interaction from a machine-centric point of view, and the processes of feed-forward to a user and feedback from the user are seen as a human-in-the-loop.

The author's interpretation of sketching as an exercise as a similar feedback loop is presented in Fig. 1.2. We may observe sketching as an exercise by dichotomising the role of an artist into an actor and a receptor, both connected to the human brain. Here, the motor skills of a user form an exponent of the actor, the sensory faculties including the visual prowess become the exponent of the receptor, and cognitive and decision making faculties are respresented by the brain. With this as the standpoint, the actor in an artist would leave an impression on the paper/medium, and the receptor in her would observe the same, and both of them coordinate with the brain. How frequently this cycle activates, and how conscious the artist is, about the decisions that have been made — are all open questions.

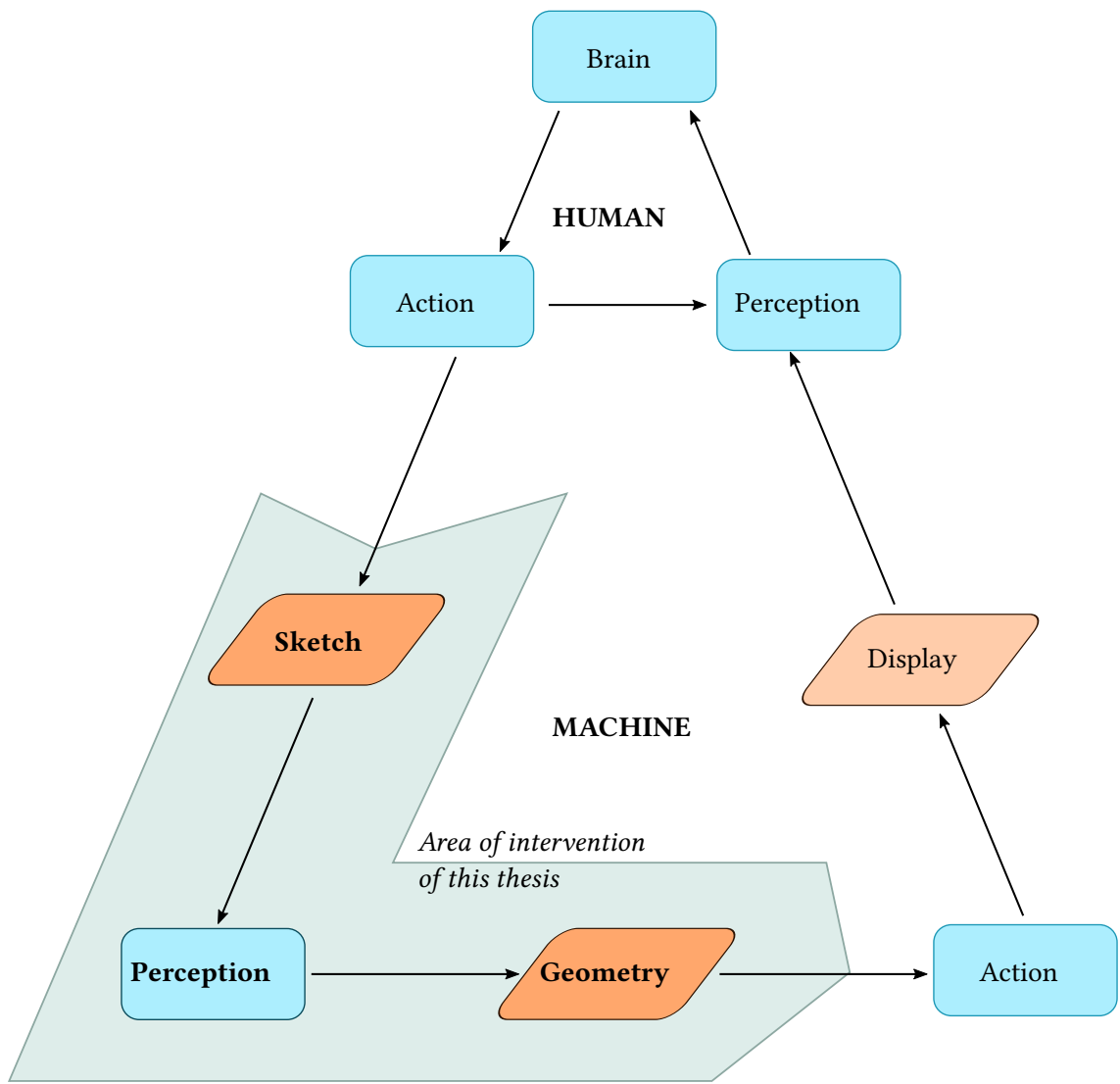

Figure 1.2: The feedback loop of sketching as an exercise.

When using automation tools, this feedback loop is extended to accommodate for the role of machines. Here the artefacts of human motor skills, eg. a sketch or a keypress, acts as a feed-forward to the machine. The artefact may, moreover, be anything ranging from a human hand movement to the body movements; motion with optical markers or special-purpose markers or motion without markers; a sketch drawn in prior or a real-time pen state on a pen-tablet or a real-time pen state in three-dimensional space. It is noteworthy that the movements, while observed by the system are also observed by the human. This thesis will limit its scope to studying pre-drawn line drawings, and further refer to it as sketches.

A machine, further processes the observations from feed-forward to deduce a machine representation based on a mathematical model, pertinent to the observations. For example, looking at the hand-written text, an OCR system would recognize the text characters word by word, line by line. The output machine representation in this case would be simple, *i.e.* the ASCII text. But there are may be many intermediate representations that may otherwise be used, *e.g.* the intermediate representations of the AlexNet and VGG model, either of them pretrained over the ImageNet dataset, has been a constant backbone for research till date. This thesis will focus on exploring mathematical models pertinent to geometric representations.

The machine representation is then processed further to be projected onto a tangible physical form and provide for a feedback from the system. A system feedback, like the feed-forward, may as well have a broad range, *e.g.* projecting on a display, a paper print, 3D printed form, a carved object using CNC-milling, a stereo-vision image for a headgear and so forth. In each case geometry may be presented as a 2D/3D wireframe, low-polygon faceted render, or a densely sampled render using a phong or a more sophisticated model. Our aim was for simple intelligible user feedback, for example we used a robot that draws in § 3 and a sketch completed with hatch patterns in § 4.

**Contribution**  Motivated by the far-reaching consequences of sketches, and with the author himself being trained architect, this thesis offers an alternate standpoint into the processing of sketches in order to distil geometric constructs that are useful to further processing downstream in a designer's workflow.

Processing of sketches from a classical standpoint have highlighted various limitations inherent to the heurism embedded in the methods to capture the order in chaos, *e.g.* a straight line drawn by every artist is different and is differently drawn every time. The details have been documented in § 2. On the other hand, with the paradigm of end-to-end training enabled by the commercial availability of advanced hardware and complementary advances in deep-learning, we witness promise in discovering a manifold that models a tight decision boundary and is yet robust to the variance of the real world.

This thesis explores the possibility of a structured interpretation with the help of deep-learning paradigm hoping to leverage the neural networks as a universal function approximator. Firstly, the § 3 attempts to infer a planar graph approximation from a sketch using the deep-learning method as simplification mechanism and feeds the output into a classical method for structured inference, further downstream in the pipeline. Secondly, encouraged by the former success, § 4 attempts to leverage the ability of deep-networks to be trained end to end, for a conditional image to image translation that requires the model to possess geometric understanding. Here the training framework was designed without explicit access to 3D information.

Other than the technical novelty, the problem identifications in this thesis follow the context of Fig. 1.2, so that the solutions once achieved may be incrementally plugged-into the workflow for subsequent evaluation by the designers with respect to its adoption and usability.

**Further in this thesis.** With the human centric feedback-loop as a background for sketching as a process, and looking into inferring geometry from sketches, this thesis shall detail out the inspirations, actions and milestones in the process of building a machine that takes human sketches as input and provides a machine representation for geometry. § 2 reviews prior approaches in literature for parsing geometry from sketches; talks about their limitations and gaps; and also briefly discusses the methodologies and recent advances that have led to problem formulation, and that have inspired the solutions presented to those problems in § 3 and § 4.

§ 3, describes the method for inference of two dimensional skeleton geometry from sketches using advances in deep learning (see § 3.3.1) followed by a classical feedback loop based approach (see § 3.3.2). It also contributes towards training of the deep-neural model by creating of a set of synthetic sketches and corresponding segmentation masks (see § 3.4.1). The utility of such

an exercise is portrayed by feeding the inferred geometry to two different drawing robots (see § 3.4.4).

§ 4 details out the attempt to complete the sketch of a three-dimensional object, using user information about illumination and style (see § 4.3). In the process, contribution was made towards usability testing of the method by designing a user interface to capture user information through the popular image manipulation tool of GIMP (see § 4.8). For the purpose of sketch completion, a single deep network was designed that captures information from three different modalities and it was trained under the generative framework to produce the completed sketch (see § 4.5). For the lack of training data pertinent to task, a large number of synthetic sketches were generated procedurally by first generating 3D solids using a solid grammar (see § 4.4.1), and then rendering it automatically using a mature graphic software (see § 4.4.3).

§ 5, concludes with a discussion on geometric inference from sketches vis-à-vis the proposed methods, their applications, limitations and future scope.

# Literature Review

"Anybody can draw" summarises the universal capacity for visual communication; be it the hieroglyphs of the primitive men or few strokes of a marker on the whiteboard of a sophisticated meeting room. Sketching is arguably a universal tool to author in a visual language. However, when it comes to modelling, sketch goes only as far as the conceptual stage. And computational tools take over as soon as a demand for high fidelity appears in the pipeline. Translating a design prototype depicted in sketches to a geometric model takes significant effort for a trained graphic artist, and often times proves to be a critical link. Scientists have hence been interested in computational methods to bridge the gap in geometric inference, and have thus given rise to its own niche in literature.

In this chapter we have described the works that relate to the problem of "geometry inferred from sketches," and in the process highlighted the research gaps, inspirations and significant contributions in literature that have guided this work. This chapter initially describes how vectorisation and segmentation have been in search of the Goldilocks balance between faithful representations and noise detection. Later, through the field of sketch based modelling and inference of 3D geometry, the problem of this thesis has been distinguished. Finally, the two problems that have been solved in the further chapters have been briefly described .

## 2.1    Vectorisation

The problem of translating a line-drawing image to a geometric representation as in vector graphics is known as vectorisation. A solution to this problem finds application in many a fields namely cartography [HKLP82; NAN88], character recognition in document parsing [PC82], technical drawing in manufacturing [CFMP84; Ble84], construction and architecture [ASAO96], and also a field as diverse as psycho-linguistics [LC85].

The early methods on line-drawing image vectorisation were based on the segmentation approach for natural images [TB77] that provided convenient primitives for processing in a downstream task. As of then, the document acquisition was primarily based on a line scanner, and a major branch of work stemmed out of run-length processing [LC85; DHTW80; Pav86]. A parallel branch may be seen as working on graphical analysis of image pixels [CFMP84; NAN88], which gave way to the application of statistical learning approaches on contemporary advances in segmentation [SM00; FH04; ASS+10; ST16].

[TB77] should be cited as one of the earliest influencing works in segmentation, albeit in the domain of natural images, and also intuitively summarise the problem of segmentation,

> *"If interpretations could be assigned to every pixel, then segmentation would be reduced to the trivial process of collecting adjacent pixels with the same labels."* [However, the challenge with pixel level segmentation is] *"the excessive volume of data and the absence of global attributes."*

The authors built upon the statistical analysis of features *e.g.* contrast, histograms *etc.*, extracted from the image regions. Later *super-pixel* based methods investigated over how to extract these regions with a common set of features, and had been the basis for both the rule based frameworks for vectorisation by segmentation and for the stochastic frameworks using energy constrained programming in the spheres of medical imaging and object detection.

### Methods based on run-length

A line-scanner would scan a single sheet document, one line at a time, looking for a mark on paper, and register the strength of such a mark as the intensity of positive signal at that point. The quality is limited by two factors, hardware resolution and noise caused by wear and tear of the hard copy. The early methods analysed the natural run length of positive signal in each line of scan to arrive at a geometrical structure.

**Thinning based**   methods found their utility in the domains of cartography [HKLP82] and character recognition [PC82], besides being used to encode human sign-language [LC85]. These methods essentially rely on encoded features obtained for example after edge detection upon a change of pixel intensity, or morphological transforms like thinning, in order to achieve their respective end, *e.g.* to define a "signature" in human sign language. It wont be far-fetched to attempt a leap-motion based gesture recognition inspired by this work.

**Decomposition based**   methods in general decompose the image into smaller syntactical entities that follow a higher order structure. In the domain of character recognition, [DHTW80] follow a syntactical approach to infer structure and counter noise using a statistical classification metric. [AS84] provide a general-purpose segmentation using a rule based classification of medial axis transform of blob shapes. In their attempt to vectorise electrical drawings, [Ble84] proposed a rule-based pattern recognition system which used the connected components of segments obtained after analysing proximity and run-lengths of a line-scanner output.

The early works also looked at graphical analysis at pixel level. In the context of engineering drawing digitisation, [CFMP84] used pixel-level connected component analysis in an image to arrive at boundary pixels in a binarised drawing. The authors further analysed the resulting boundary graphs for presence of hierarchical grammatical patterns to simplify the structure and eliminate noise. Graphical analysis of extracted contours using a pairing relationship was shown to satisfactorily generate skeleton geometry for engineering drawings [HF94]. In the context of document analysis, a graphical analysis over inferred sketch primitives, namely line segments and closed curves, has been documented as faithful representation for downstream tasks [RVE00].

### Curve fitting

While earlier attempts at vectorization and skeletonisation use classical vision based strategies of thinning and contour identification to simplify the image, hardware capabilities had advanced by the turn of the century and attempts were documented towards vector inference using line fitting or polygon fitting [Tom98; TT00; Doe98]. These may be classified broadly into two branches namely, *a)* the skeleton based methods, that dealt with corner artefacts; and *b)* the contour based methods, that dealt with problems related to one-to-one correspondences.

**Observation.**   It is observed that the problem of interpreting sketches as a geometry has not been framed as interpreting a planar graph from a rasterised line drawing. Fundamentally, a graph represents a superior structural richness as compared to a sequence, which forms our standpoint in the first investigation (see § 3). The early attempts at vectorisation perform well with binarised images, and inspire us towards solving the fundamental problem, but also show that noise is a major deterrent towards their performance. Recent methods both analytical and statistical in nature, are more robust to noise but invariably these methods treat sketches as a sequence of strokes. The recent success in image segmentation based on neural processing lies either in the domain of natural images or they target a higher level problem of semantics. The gap identified here, is arguably located at a much lower level of distinguishing corners from lines. This thesis draws inspiration from the early methods where a high fidelity segmentation forms the basis for syntactic inference, and in the same spirit presents as review of literature in segmentation and labelling of images in the following section § 2.2.

## 2.2   Segmentation and labelling

This section presents an appreciation of advances segmentation which has provided the basic framework for most of the works in vectorisation as well as for sketch based modelling (See 2.5) With a super-pixel based approach, the methods result in processable primitives that are far less in number as compared to the number of raw pixels in an image. Further there had been advances in segmentation of sketch parts based on a probabilistic framework. In the latter half of last decade, the advances in neural computing based approaches have registered significant progress. A brief review of each has been covered in the following subheads.

### Super-pixel based approach

Earlier pixel-level approaches, had suffered from two shortcomings, namely *a)* computational complexity, which was often times countered by using downsampled images as input, and hence led to inferior evidence of model performance; and *b)* a highly localised inference, with little or no contribution from higher order image structure. With advances in hardware and processing capacity, the processing principle showcased by [TB77], that "picture elements may be clustered based on semantic interpretations" gave way to super pixel based methods dominating the segmentation spectrum in the latter part of first decade of the new millenium.

The underlying methods explored strategies for clustering the pixels as super-pixels to create a higher order planar graph, consisting of super-pixels as vertices with cardinality reduced by orders of magnitude compared to the count of original pixels. For example, an image with a million pixels, should be reduced to one having 16k super-pixels ~ 64 px in size; or 4k super-pixels ~ 256 px in size; or 1k super-pixels ~ 1024 px in size. The super pixels provide a three fold advantage: a) provide for convenient local primitives to compute features; b) capture redundancy at local level; and c) drastically reduce the downstream complexity.

Several algorithms had been proposed in this spirit, broadly classifiable as either graph-based methods or as gradient-ascent based methods. The graph-based algorithms work on defining cuts to create partitions in an image pixel graph, e.g. a method inspired by perceptual grouping [SM00], human-annotated data-driven approach to segment natural images [RM03], and their use to predict human body pose [Mor05].

Into the family of graph-theoretic approaches, Felzenszwalb *et al.* [FH04] contributed with a fundamental algorithm based on Kruskal's algorithm for finding minimum spanning tree [CLRS09] but it assumes no priors and yet, it efficiently computes the super-pixels in a time complexity log-linear in the count of super-pixels. This method is independent of the choice of *similarity measure* used to define inter-cluster relationships, and it has thus, been used in the seminal super-pixel based segmentation [ASS+10], and generic methods for scene parsing like [TL10], and inspired methods for diverse fields like object recognition [UvdSGS13]. Energy optimisation framework with pixel-level energy definition had also been used to formulate the super pixel estimation, and solved using approximate graph cuts [VBM10].

The gradient-based algorithms start with a rough clustering and rely on the definition of a gradient to refine the clusters in each successive iteration eventually, until convergence using methods often-times specific to the gradient-definition itself. Broadly speaking, the definition of cluster is dependent on a representative point called seed, for example a centroid [CM02; VS08; FS12] which has an associated manifold to ascend along the gradient definition. Variations for example the watershed based method [VS91] are seen in algorithm where ascent is performed from a local minimum in the image plane to the ridge lines of watershed. Turbo-pixels [LSK+09] utilise local image gradient to define the geometric flow of super-pixel seeds. In SLIC [ASS+10] the authors use a linear combination of pixel intensity difference in Lab color space and the euclidean distance between the seeds to define the gradient.

**Probabilistic models**

Relatively recently, the probabilistic models found application in segmentation and labelling of sketches in graphics and vision. Schneider *et al.* [ST14] had successfully used an ensemble of Gaussian Mixture Models and Support Vector Machines with Fisher Vectors as distance metric to discriminate strokes; Further, it was illustrated that use of Conditional Random Fields over a relationship graph described by proximity and enclosure between the strokes, performs well in assigning semantic labels to strokes [ST16]. The classical methods use hard coded features like SIFT [Low04], that was used here, as the basic building block for learning algorithms.

**Neural processing methods**

Recently, convolutional neural networks have been proven to be highly effective in classifying segmented strokes as given labels [ZXZ18]. Li *et al.* [LFT18] have illustrated the effectiveness of U-Net based deep neural network architecture in the context of semantic segmentation, with the help of 3D objects dataset. The neural methods also exploit the characteristics of dataset, to inherit a *look-ahead bias*, for example data synthesised out of 3D objects, used by Li *et al.* [LFT18], have a strong sense of enclosure, which are not necessarily true for the hand drawn counterparts.

Methods, in the recent literature of deep-learning, have shown success in semantic segmentation of sketches and their generation using sequence-to-sequence translation. These methods have been shown to be successfully deployed both as a discriminative model [WQLY18] as well as a generative model in the context of sketch segmentation [KS19]. Such sketches, however, have been studied as a sequence of strokes, such that each stroke is described as an ordered set of line segments, represented as vector geometry. A pen lift results in the end of a stroke data and a subsequent pen-touch on canvas marks the start of next stroke data. As naturally follows, the generative model [KS19] has been used to reconstruct a sketched symbol. The success here also encourages to use segmentation using deep networks over sparse spatial signals like the sketches itself, to obtain a representation useful for further downstream tasks.

**Favreau's observation**    Recent developments in vectorisation deal with noise removal and Favreau *et al.* [FLB16] in particular, review the methods the trade off between fidelity vs simplicity, a question that embodies the classical question of variance vs bias in statistical learning. Although the authors provide the users with an affordance to control the output and find their Goldilocks balance, loss of information vis-à-vis image domain is pretty apparent.

## 2.3    Robots that draw

Research in robotics has seen some recent developments, while trying to create a robot that draws. Paul, the robotic arm, was trained to sketch a portrait, with user input, while deploying a stack of classical vision methods into their pipeline [TF13]. More recently, [GK17; GKAK16] investigated whether a similar feat can be achieved through a quad rotor, aiming "to apply ink to paper with aerial robots." Their contributions include computing a stipple pattern for an image, a greedy path planning, a model for strategically replacing ink, and a technique to adjust future stipples dynamically based on past errors. Recently, robots with high quality manipulators have been used to draw line based art on uneven surface(s), exploiting the availability of impedance control [SLK18]. Earlier, Fu *et al.* [FZLM11] illustrated an effective use of graph minimisation over hand crafted features to predict a reasonable ordering of given strokes. Primarily, such efforts are based on space filling exercise, and focus on the control systems, rather than the accuracy of the input image. *This inspires us to develop a feedback-loop as a principle for vectorization in our investigation, although we capture the line nature of artwork, and it is a different exercise altogether.*

## 2.4    Sketch Dataset

With the advances in neural processing techniques, it is imperative to review the availability of large datasets pertinent to a given task, because it would not be an over estimation for a neural model to consist of tens of millions of floating point parameters, and thus demand large amount of samples. The strategy to acquire such data has been documented as having two traits, *a)* one, that of being playful to a contributor, *i.e.* either look at and redraw or draw from memory, but without fearing failure or aiming for perfection; and *b)* two, to enforce a time constraint, say less than half a minute. We had seen the use of Huang *et al.*'s dataset with 10 classes of 30 examples each [HFL14], and the TU Berlin dataset, by Mathias *et al.*, having 20k drawings split into 250 classes [EHA12]. However, the contribution of the *Quick-draw* dataset [HE17] with ~50M samples split into 345 classes, was a boon for neural methods. Since then, we have seen many sketch-based investigation, *e.g.* stroke segmentation [WQLY18], sequential generation of sketch strokes [KS19], sketch completion [CCSY19], and so forth aided by the exhaustive dataset. *It follows that Quick-draw is a natural choice of experimental data for our sketch based investigation.*

## 2.5   Sketch-based modelling

**Interactive approach.**   The pioneers, here should be mentioned starting with SketchPad [Sut64] by Sutherland; Fuzzy Spline Identifier [Sag95]; the 3D sketching interface SKETCH [ZHH96] by Zeleznik *et al.*, which forms the basis for many a future sophisticated modelling interfaces like SketchUp [SGI00]; and Igarashi's seminal GIGA [KIM96], and Teddy [IMT99] that inspired many a following interactive modelling tools in computer graphics like FiberMesh [NISA07] Plushie [MI07] and SketchChair [SLMI10]. The latter works allow a user to express a 3D shape using a 2D drawing, but naturally, using a few priors, and focus on interactivity, since "image at the eye has countless interpretations" [Hof00]. It is worthy of mention that SKETCH [ZHH96] aims to provide enough affordance to combine two-dimensional shapes to create a 3D solid primitive, for example using extrusion, revolution, slicing, curve-following and so forth essential to create an hierarchical structure like the constructive solid geometry.

Igarashi's Teddy [IMT99] and its relatives focus on gestural interaction to create rotund 3D objects using surface models, and refine them. GIGA [KIM96] offered production level interactivity with gestural communication, which have carried over to date, for example scrubbing over a segment to indicate erasure. Teddy's [IMT99] modelling techniques are a reinterpretation of implicit functions [BW90] (from the Xerox Lab) as low polygon models using the skin algorithm [MCCH99] that iteratively evolves the triangle mesh after repositioning particles and redefining connectivity until convergence. Plushie [MI07] improves over Teddy with a practical utility to create stuffed toys providing for seams automatically. Similarly, FiberMesh [NISA07] improves over Teddy, in the geometric interpretation of curves using differential coordinates and constrained co-rotations; and the interpretation of surfaces by constraint programming for surface normals using partial differential equations (PDEs) approximated by a least squares minimisation problem. Additionally, FiberMesh allows for modification in mesh topology, albeit with a noticable overhead, and thus it is invoked only on demand. In another work [RDI10], the authors propose to use sketches as input for a CSG based model. The method follows a simple and intuitive principle to intersect the extruded silhouettes of orthographic projections of an object, input as a sketch and interpreted as a geometric curve. Any further details, for example concavity or depth of the container may be modelled as a CSG and assembled. More recently, SMARTCANVAS [ZLDM16] extended the works relying on planar strokes.

**Analytical Inference,** for 3D reconstruction is another popular approach to modeling from sketches. With the context of vector line art, [MM89] inferred simple curved shapes, and [LS96] showed progress with complex compositions of planar surfaces. Recently, for illuminating sketches, [SBSS12; IBB15] had shown the effectiveness of normal-field inference through regularity cues, while [XGS15] used isophotes to infer the normal-field.

**Deep Learning Techniques,** have more recently, shown substantial progress in inferring 3D models from sketches: a deep regression network was used to infer the parameters of a bilinear morph over a face mesh [HGY17] training over the *FaceWarehouse* dataset [CWZ+14]; a generative encoder-decoder framework was used to infer a fixed-size point-cloud [FSG17] from a real world photograph using a distance metric, trained over *ShapeNet* dataset [CFG+15]; U-Net-like networks [RFB15] were proven effective, to predict a volumetric model from a single and multiple views [DAI+18] using a diverse dataset.

**Observation.** 2D to 3D is a problem ill-posed in general sense, but the methods mostly hope to solve a subset of the problem inspired by human visual intelligence; *e.g.* point-cloud representation for cars [FSG17], signed distance functions for faces [HGY17] and so forth. Deep volumetric inference [DAI+18] attempted a more general purpose inference, but suffers from artefacts of low resolution in geometry space, which it partially mitigates with normal estimation [DCLB19]. This had inspired us to look into a method that approximates the sketch inference followed by a sketch-like render, *i.e.* to generate hatch pattern for a sketch given the lighting condition (§ 4). Such an insight leads us to review the image to image translation methods. (see § 2.6)

## 2.6 Stylisation

Thus far this chapter has described the discriminative methods of geometry inference from sketches. From a utilitarian standpoint, an inferred geometry may at an instant only be seen from a certain camera pose, which in itself is an image. This section looks at the process starting from inference upto rendering as a whole pipeline, and appreciates relevant literature from a generative point of view. Contextually put, Hagen said [MFB+88],

> "The goal of effective representational image making, whether you paint in oil or in numbers, is to select and manipulate visual information in order to direct the viewer's attention and determine the viewer's perception."

In the same spirit, early works [Hae90; SABS94; CAS+97] focused on attribute sampling from user input to recreate the image with arguably a different visual language and accentuation and offer interactivity as a medium of control. Thereby followed, a shift towards higher order controls, to achieve finer details, for example temporal coherence [Lit97], modelling brush strokes [Her98], and space-filling [SY00]. Space-filling had been buttressed, on one hand with visual-fixation data and visual-acuity [DS02; SD02], and on the other with semantic units inferred using classical vision techniques [ZZXZ09]. The finer levels of control, in these works, allowed for interactivity with details in the synthesised image.

**Image Analogies**   had popularised the machine learning framework for stylising images, with analogous reasoning $A : A' :: B : B'$, to synthesise $B'$, using image matching [HJO+01]; to achieve temporal coherence [BCK+13]; by synthesising the best-matching features retrieved from images [FJS+17; FJL+16; JFA+15]. The problems thus formulated, had allowed the use of high level information as a reference to a style, for example a Picasso's painting.

**Deep Learning Techniques**   have recently documented success in similar context, using the Conditional Generative Adversarial Networks (CGAN) [MO14] framework combined with sparse input [IZZE16]; combined with categorical data [WLZ+17]; or with an aim of *cartoonization* of natural images [CLL18].

Researchers have recently contributed to line drawings problem pertaining to sketch simplification [SISI16; SII18a]; to line drawing colorization [FHOO17; KJPY19; ZLW+18]; and to line stylization [LFH+19]. Sketch simplification [SISI16; SII18a] aims to clean up rough sketches by removing redundant lines and connecting irregular lines. Researchers take a step ahead to develop a tool [SII18b] to improve upon sketch simplification in real-time by incorporating user input. It facilitates the users to draw strokes indicating where they want to add or remove lines, and then the model will output a simplified sketch. *Tag2Pix* [KJPY19] aims to use GANs based architecture to colorize line drawing. *Im2pencil* [LFH+19] introduce a two-stream deep learning model to transform the line drawings to pencil drawings. Among the relevant literature we studied, a recent work [ZLB20] seems to be most related to our approach, where authors propose a method to generate detailed and accurate artistic shadows from pairs of line drawing sketches and lighting directions.

In general, it has been argued that models over the deep learning paradigm handle more complex variations, but they also require large training data.

## 2.7   Relevance of deep learning

This chapter has highlighted the community's persistent efforts since the eighties to solve the problem of structural inference through classical paradigm of vision. Most of the early methods were not scalable because of impertinent fringe artefacts, eg. in case of curve fitting to line-drawings [TT00; Tom98; Doe98], problems persisted as corners in case of thinning, and as correspondences in case of contour-following. More recently, [FLB16] showed that sketches are predominantly high-variance, and even the best methods present a trade-off between simplicity of geometry-representation and its accuracy.

Coupled with advances in commercial hardware availability, deep learning in contrast, have promised robustness against the high-variance of real-world data, and hence a formidable candidate to infer a robust model.

## 2.8   Summary

This chapter has described historical approaches to, as well as state-of-the-art for, sketch vectorisation (§ 2.1) and sketch-based modelling (§ 2.5). The crucial observations in these relations have been documented for both the approaches. The observations point toward another related fields, which have also been thoroughly reviewed here.

The next two chapters have detailed out the context for each of the problems, namely *a)* can i teach a robot to replicate a line drawing (§ 3), and *b)* shad3s: a model to sketch shade and shadow (§ 4); have proposed a method to solve them, and have documented the experimental setup and results for each.

CHAPTER 3

# 2D Prediction

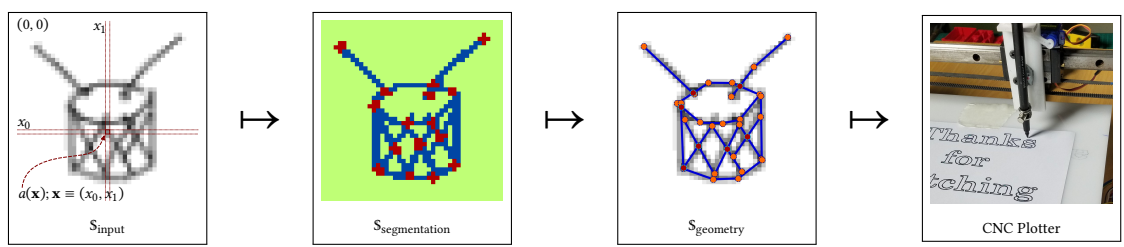

This chapter deals with interpretation of a 2D-line drawing as a graph structure, and further illustrates its effectiveness through its physical reconstruction by a robot[1]. We setup a three-stage pipeline to solve the problem as shown in the figure above. Firstly, we compute the segmentation map of the input sketch image. We achieve this using a combination of deep convolutional neural networks learned under a supervised setting for pixel-level estimation. Secondly, we estimate the the geometry as a skeleton graph. The vertices of the graph are estimated from the segmentation mask with sub-pixel level accuracy. And the edges are estimated using a feedback-loop-based setup by comparing against the input image. Finally, we perform data-interchange to a robot legible ASCII format, and thus teach a robot to replicate a line art.

## 3.1   Introduction

Line is a fundamental building block in art form, and so has been emphasised by pioneers of historical art movements led by the likes of Picasso, Paul Klee, Kandinsky and Piet Mondrian in the past. In modern times drawing as an activity has been studied in the context of problem solving in a design process, while sampling mostly over line art [UWC90; Vis06].

There is one question that forms the under-current of our investigation here,

> If I provide *any* line drawing to a robot, how can I teach it to *look at it,* and draw
> the same on a paper using a pen.

This is a different challenge that has to the best of our knowledge not been addressed previously. There are vectorization based methods in Computer Graphics that aim to convert an input sketch into strokes. However, these would lose salient stroke information and would not preserve the line drawing (refer § 3.4.5). Other approaches aim to parse sketch into semantic segments. Again, this would not be useful for obtaining stroke information. In contrast to these, we aim to *replicate* the line drawing by converting an input raster line drawing image into a set of strokes (refer § 3.3) and the corresponding instructions (refer § 3.3.3) that can be used by an automated system. To obtain this, we present an approach different from the popular vectorization models [FLB16; TT00], the segmentation models [ST16; KS19], the interesting developments of the drawing robots [GK17; TF13] and further present its applicability.

We treat this problem as a special case of vectorization to infer a skeleton graph structure, by learning a function to map a line drawing from image domain to a skeleton graph structure. To this effect, we first perform an image-to-image translation, by deploying a novel *U-Net [RFB15] based deep neural network* (DNN) and segment the pixels in the input sketch image as one of: a *vertex*, an *edge* or the *background* (refer § 3.3.1). The next step is to infer the graph structure from the segmentation mask in two steps, vertex inference and the edge inference respectively. Later, this is further translated to GCODE to input to a CNC plotter, or used directly by factory software to compute trajectory for a robotic arm. The pipeline in further detailed out in § 3.3.

Deep neural networks, in a supervised-setting, demand a formidable size of labeled dataset at training stage. For this task of segmentation, we propose a novel dataset built upon the *Quick-draw* dataset [HE17], to generate the segmentation true labels, on the fly. The *Quick-draw* provides a sequence of strokes for each drawing, which we rasterize using a trivial OpenCV-based rasterizer, and further compute the segmentation true labels on the fly. This may be treated as an *entry-point* for adapting the method for more sophisticated use-cases, by generating a dataset using an alternative rasterizer. The dataset creation is further detailed out in § 3.4.1.

Given an input with a segmentation mask, we perform graph inference (or vectorization) using a simple yet effective method based on classical vision and learning and infer the geometry as a graph data-structure, from the sketch input in image domain. We utilise the segmentation of the image into *a) vertex channel:* where for most practical cases the vertex-islands have one to one correspondence with vertices of the graph; and *b) edge channel:* that shows a similar behaviour for edges. The inference is made using an iterative update to a parametrized filter criterion, that selects a subset of the edges as graph proposal after every update. Graph inference is further detailed out in § 3.3.2.

To extract stroke sequences from the graph structure, we effectively utilise a recursive approach and further illustrate its utility using two devices, namely CNC plotter, and a robotic arm (see § 3.3.3).

In this chapter we make highlight the main contributions as follows:

- Propose a multistage pipeline, for segmentation, and graph interpretation;
- Novel architecture for the deep segmentation network;
- Annotated dataset for training the network;
- Improvised loss function as the training objective; and
- Feedback based iterative update to infer graph structure from image domain.

## 3.2 Relevant Works

**Sketch vectorization** is a historically studied problem, with early attempts utilizing classical vision based methods, and later attempts using probabilistic methods followed by a seminal work highlighting the pivotal trade-off between accuracy of captured information and simplicity of the inferred model [FLB16]. Earliest attempts at vectorization and skeletonization use classical vision based strategies of thinning and contour identification to simplify the image, followed by vector inference using line fitting or polygon fitting [Tom98; TT00; Doe98]. Two specific approaches include skeleton based methods, that have to deal with corner artefacts, and contour based methods, that deal with problems related to one-to-one correspondences. Recent developments deal with noise removal and Favreau et.al. [FLB16] in particular, investigate a higher order problem of fidelity vs simplicity of the method. *Inherent to these approaches is the* *loss of information from the image domain in order to mitigate the basic problems.* We propose to use segmentation techniques which are comparatively better at preserving the information as shown further in § 3.4.5.

**Segmentation and labelling**   of sketches has been subject to profound research in graphics and vision. Schneider *et al*. [ST14] had successfully used an ensemble of Gaussian Mixture Models and Support Vector Machines with Fisher Vectors as distance metric to discriminate strokes; Further, it was illustrated that use of Conditional Random Fields over a relationship graph described by proximity and enclosure between the strokes, performs well in assigning semantic labels to strokes [ST16]. The classical methods use hard coded features like SIFT, that was used here, as the basic building block for learning algorithms.

Recently, convolutional neural networks have been proven to be highly effective in classifying segmented strokes as given labels [ZXZ18]. Li *et al*. [LFT18] have illustrated the effectiveness of U-Net based deep neural network architecture in the context of semantic segmentation, with the help of 3D objects dataset. *The problems involving semantic segmentation whether used in classical context or with the deep learning based methods, target a higher level problem, namely semantics. We formulate our problem at a much lower level, by distinguishing corners from lines.* The neural methods also exploit the characteristics of dataset, to inherit a *look-ahead bias*, for example data synthesised out of 3D objects, used by Li *et al*. [LFT18], has a much stronger sense of enclosure, which are not necessarily true for the hand drawn counterparts, that form our dataset.

Methods, in the recent literature of deep-learning, have shown success in semantic segmentation of sketches and their generation using sequence-to-sequence translation. These methods have been shown to be successfully deployed both as a discriminative model [WQLY18] as well as a generative model in the context of sketch segmentation [KS19]. Such sketches, however, have been studied as a sequence of strokes, such that each stroke is described as an array of continuous line segments, represented as vector geometry. A pen lift results in the end of a stroke data and a subsequent pen-touch on canvas marks the start of next stroke data. As naturally follows, the generative model [KS19] has been used to reconstruct a sketched symbol. *Although sequences do make sense to work with, in stroke domain, it implies that a sequence of vector input is available to begin with; whereas, our work deals with a raster image as an input.* The success also encourages to use segmentation using deep networks over sparse spatial signals like the sketches itself, to obtain a representation useful for further downstream tasks.

**Research in robotics** has seen some recent developments, while trying to create a drawing robot. Paul, the robotic arm, was trained to sketch a portrait, with user input, while deploying a stack of classical vision methods into their pipeline [TF13]. More recently, [GK17; GKAK16] investigated whether a similar feat can be achieved through a quad rotor, with the aim of "applying ink to paper with aerial robots." Their contributions include computing a stipple pattern for an image, a greedy path planning, a model for strategically replacing ink, and a technique for dynamically adjusting future stipples based on past errors. Primarily, such efforts are based on space filling exercise, and focus on the control systems, rather than the accuracy of the input image. This inspires us to develop a feedback-loop as a principle for vectorization in our investigation, although it deals with capturing the line nature of artwork, and is a different exercise altogether.

Robots with high quality manipulators [SLK18] have been used to draw line based art on uneven surface(s), exploiting the availability of impedance control. Earlier, Fu et al. [FZLM11] illustrated an effective use of graph minimisation over hand crafted features to predict a reasonable ordering of given strokes. However, these methods rely on vector input. We on the other hand propose a pipeline that reads from a grayscale image.

**Among the datasets,** since the contribution of the *Quick-draw* dataset [HE17], we have seen many sketch-based investigation in the community, e.g. segmentation [WQLY18], sequence generation [KS19], sketch completion [CCSY19], and so forth, where the dataset provides for an exhaustive sketch prototype.

## 3.3 Methodology

The problem is decomposed into three parts, forming the pipeline as shown in Fig. 3.1, namely segmentation, graph inference, and translation. The first part takes the raw sketch, a grayscale image as an input, and translates it into a segmentation mask, one for each label, which in our case are *background, vertices,* and *edges.* The second part uses the segmentation channels of lines and corners as input and infers a graph structure, so that its vertices represent the cusp or end points, and its edges represent the curve drawn between them. The last part takes the graph structure and partitions it into a sequence of strokes using a recursive approach.

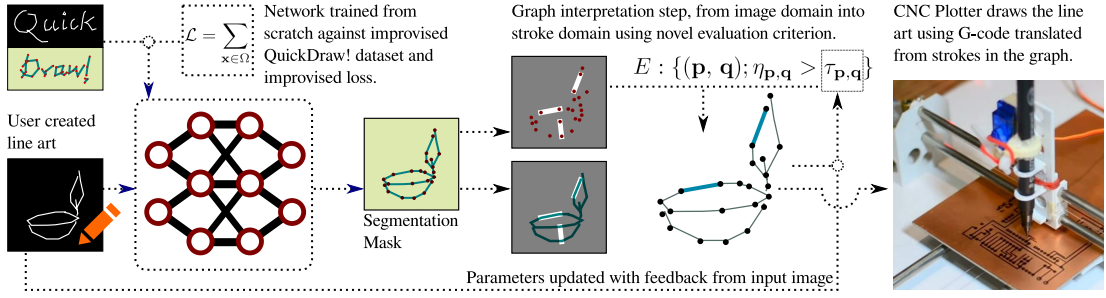

Figure 3.1: Overview of 2D prediction method. *Left to Right*: An artist drawing; digital copy of the drawing; segmented version; drawing by a CNC plotter.

### 3.3.1   Segmentation

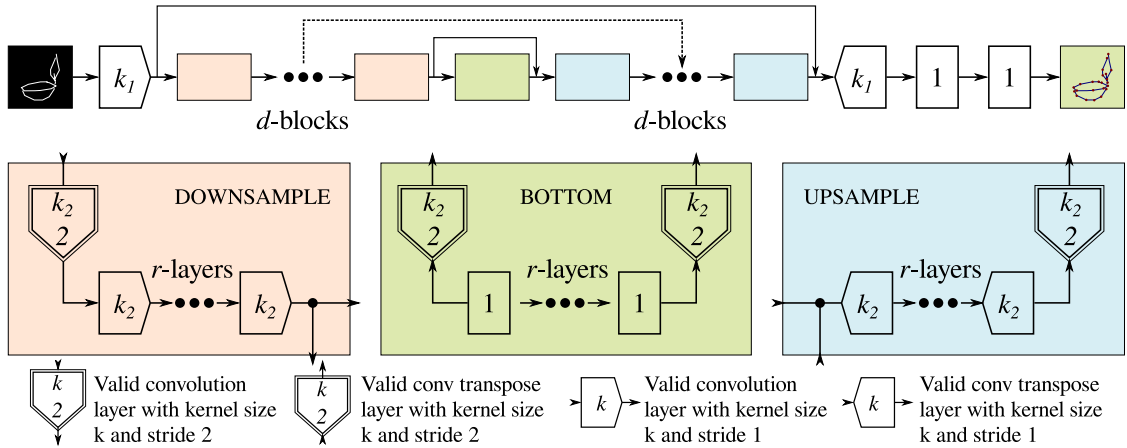

Figure 3.2: Reinterpretation of the U-Net architecture with modular blocks for *downsampling*, *upsampling* and *bottom*.

We intend to learn a segmentation function, $\mathscr{F}(\mathbf{x}, I; \theta)$ parameterized by $\theta$, that given an image $I$ and pixel location $\mathbf{x} \in \Omega \subset \mathbb{R}^2$ in spatial limits of image, returns a factor representing probability $\mathbf{p} \in \mathbb{R}^k$ of $\mathbf{x}$ belonging to each one of the $k = 3$ classes, i.e. $p_k$. This is implemented using a DNN, a variant of U-Net, that avoids cropping the images, and retains the valid convolution operations.

U-Net, a variant of deep neural encoder-decoder architecture with skip connections, was proposed by Ronneberger [RFB15] and was an extension of fully convolutional network (FCN) [SLD17; LSD15].It was shown to be highly effective in the context of biomedical segmentation. Compared to the FCN, U-Net proposed a symmetric architecture, to propagate the skip connections, within the convolutional network. Its effectiveness in the context of semantic segmentation was also shown by Li *et al.* [LFT18].

Additionally, since downsample/upsample is followed by valid convolutions, U-Net forwards only a crop of the activations as a skip connection. Consequently, the output of the vanilla U-Net produces an output image with size smaller than that of the input image, $H_{\text{out}}, W_{\text{out}} < H_{\text{in}}, W_{\text{in}}$. As a counter measure, the training and test images were augmented by tiled repetition at borders, for the vanilla U-Net.

We on the other hand create an exact mirror of operations unlike the U-Net architecture, i.e. downsampling is followed by valid convolutions and upsampling follows valid transpose convolutions. This eliminates the need for a tiling-based augmentation, and further gives way to nice convolution arithmetic.

Yu et.al [YYL+17] observed that in the context of sketches, the network is highly sensitive to the size of "first layer filters", the "larger" the better. Although they used a kernel size of ($15\times15$) for the first layer, we found that a size as low as $7 \times 7$ is effective in our case.

Another notable difference, inspired by Radford et. al's observations [RMC15], has been in the downsampling and upsampling operations. U-Net uses "$2\times2$ *max pooling operation with stride 2 for downsampling*" [AND *a corresponding upsampling strategy.*] We have, on the other hand, used valid strided convolutions for downlasmpling, and valid strided transpose convolutions for up-sampling.

Our network, as shown in the Fig. 3.2 is parameterised with a 4-tuple,$(k_1\, k_2\, d\, r)$, where

- $k_1$ is the kernel ($k_1 \times k_1$) of the first convolution layer.
- $k_2$ is the common kernel ($k_2 \times k_2$) of all subsequent layers.
- $d$ is the depth of the encoder/decoder, that is to say number of downsampling operations in the encoder or upsampling operations in the decoder .
- $r$ is the number of convolution operations along each depth level.

### Training Objective

We started with the standard, weighted softmax cross entropy loss, as in the case of U-Net. For each pixel in the image, $\mathbf{x} \in \Omega; \Omega \subset \mathbb{Z}^2$, if the network activation corresponding to a given label $i$, be expressed as, $a_i(\mathbf{x});\ 0 < i \leqslant K$, $K$ being the number of labels, we express the predicted probability of the label $i$, as the softmax of the activation, $p_i(\mathbf{x}) = \exp(a_i(\mathbf{x}))/\sum_j^K \exp(a_j(\mathbf{x}))$. We want $p_{\ell(\mathbf{x})} \approx 1$ and $p_i(\mathbf{x}) \approx 0; i \neq \ell(\mathbf{x})$ for the true label $\ell(\mathbf{x})$. So we penalise it against the cross entropy, weighted possibly at per pixel level $w(\mathbf{x})$, as follows,

$$\mathscr{L} = \sum_{\mathbf{x} \in \Omega} w(\mathbf{x}) \log(p_{\ell(\mathbf{x})}(\mathbf{x})) \tag{3.1}$$

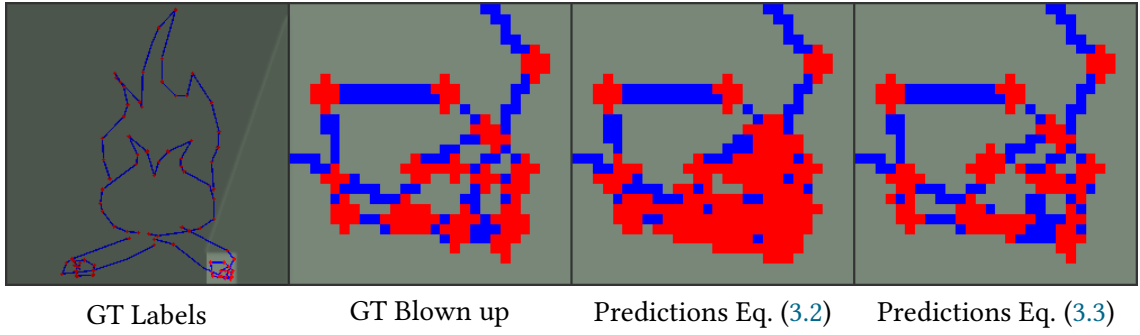

| GT Labels | GT Blown up | Predictions Eq. (3.2) | Predictions Eq. (3.3) |

Figure 3.3: Effect of the change in weights for cross entropy loss. *Left to Right.* True labels highlighting selected area; Blown up version of selected area in true labels; Prediction in selected area using standard weights as in Eq. (3.2); Prediction in the selected area using max weights as in Eq. (3.3).

**Class Balancing**   is a standard technique in machine learning based approaches and noticeably improves the performance. *Consider a task to choose a ball at random from a collection of 100 balls which may be red or black, wherein without looking I predict the colour to be red. Now suppose, the collection consists of only one black ball and the rest all as red. I would be 99% correct! The focus should thus shift towards predicting the black ball correctly.*

To encourage learning, as per the class balancing method we weigh the cost of predictions by inverse (reciprocals) of class probabilities, *i.e.* in Eq. (3.1),

$$w(\mathbf{x}) = \omega(\ell(x)) = \frac{\dfrac{1}{p_c(\ell(\mathbf{x}))}}{\sum\limits_{l \in \text{labels}} \dfrac{1}{p_c(l)}} \tag{3.2}$$

In our experiments this was already promising as shown in the third image from left in Fig. 3.3. But we may also see that closely spaced vertices are estimated to be a large blob. To further improve our results, we investigated a variant using max weights,

$$w(\mathbf{x}) = \max(\omega(\ell(\mathbf{x})), \omega(\rho(\mathbf{x}))) \tag{3.3}$$

where $\rho(\mathbf{x})$ represents the predicted label. This significantly improves the result, as shown in the rightmost image in Fig. 3.3. Even the prediction of very complicated structures resembles the true distribution.

### 3.3.2    Graph Interpretation

At this stage of the pipeline, we have the input image $I$, and the segmentation masks $M_{\text{vertex}}, M_{\text{edge}}$, as the information available to us. And, we aim to obtain the vector representation of vertices, $V \equiv \{\mathbf{x} : \mathbf{x} \in \mathbb{R}^2\}$ and edges $E \equiv \{(u, v) : u, v \in V\}$, to obtain the graph $G(V, E)$.

We make two observations in this regard. The first observation is regarding the vertex segmentation mask that helps us infer the vertices with a sub-pixel level accuracy. The vertex mask $M_{\text{vertex}}$, is a binary spatial signal so that only the pixel neighbourhood of a vertex have a positive response. This results in disjoint signal islands, say $N$ in number. We detect these islands, using the standard connected component analysis, each as a separate set of pixels, $X_j; 0 < j \leqslant N$. The centroids of each set give us the vertices $V$ as in (3.4).

The second observation is that in the edge segmentation mask, for any edge $(u, v) \in E$, there is a strong response over a thin rectangular ROI along the line connecting $u$ to $v$. In contrast, the response is weak for any pair of vertices $(u, v) \notin E$. More formally, let $M_{(u,v)}^{\text{ROI}(\beta)}$ be a thin rectangular ROI of width $\beta$ aligned and centered along the line $(u, v)$; and let its response be $\eta_\beta(u, v)$ as in (3.5); then the edges $E$ may be obtained as in (3.6).

$$V \equiv \{\text{centroid}(X_j) : 0 < j \leqslant N\} \tag{3.4}$$

$$\eta_\beta(u, v) = \mathbb{E}\left[ M_{\text{edge}}(\mathbf{x}) \mid \mathbf{x} \in M_{(u,v)}^{\text{ROI}(\beta)} \right], \tag{3.5}$$

$$E(\beta, \tau) \equiv \{(u, v) : \eta_\beta(u, v) > \tau, (u, v) \in V, u \neq v\} \tag{3.6}$$

where $\mathbb{E}[\cdot]$ is the expected value operator; and $\tau$ representing threshold and $\beta$ representing rectangular mask width are model hyper-parameters. It is noteworthy that we have assumed so far that such a threshold $\tau$ exists, which can discriminate between an edge and a non-edge pair of vertices using the ROI response as a measure. We have experimentally verified using the formulation in (3.7) that our assumption is indeed fair.

Rephrasing our assumption, the correctness of our method rests on the premise that there exists a comfortable gap between the response values of pairs of vertices representing edges, and the pairs without edges. Formally put, $\exists\ \epsilon > 0$ such that the gap,

$$\gamma(\beta) = \tau(\beta) - \mathrm{E}\left[\eta_\beta(p, q) \mid p, q \in V\right] > \epsilon \tag{3.7}$$

$$\tau(\beta) = \min_{(u,v) \in E} \eta_\beta(u, v)$$

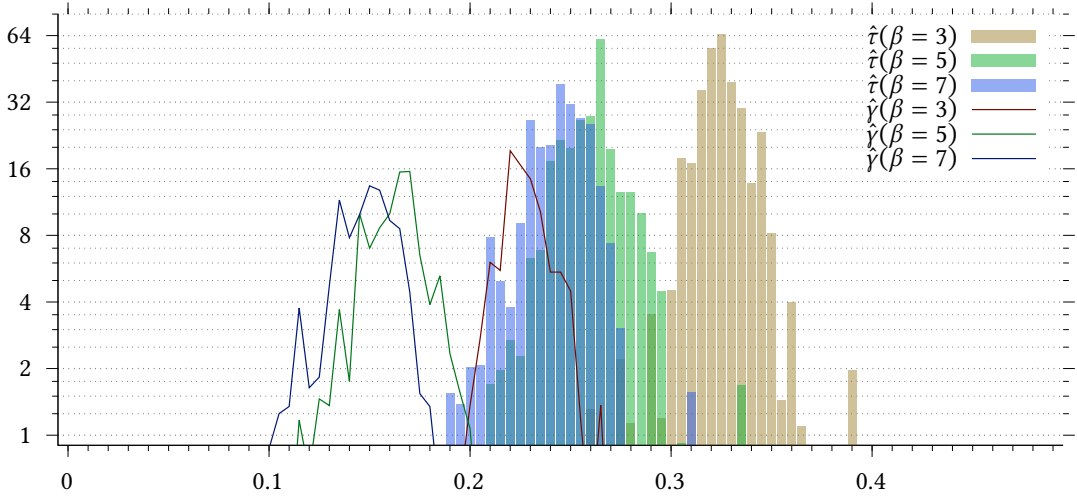

Figure 3.4: Logscale histogram of threshold $\hat{\tau}(\beta)$ and gap $\hat{\gamma}(\beta)$ approximated using $\beta \in \{3, 5, 7\}$.

In order to validate our claim we randomly selected 1024 samples from a subset of the *Quick-draw* dataset as synthesised in § 3.4.1, containing the source images, and their corresponding segmentation masks for each sample. We further approximated their threshold values $\hat{\tau}(\beta)$ and gaps $\hat{\gamma}(\beta)$ as follows. For a given drawing, we know the ground truth edge segmentation mask and the number of edges $m$, and the set of vertices $V; |V| = n$ from the dataset. For each pair of distinct vertices, we compute their responses $\eta_\beta(u, v); u, v \in V, u \neq v$ using (3.5) and arrange them in descending order $\{\eta_\beta^{(1)}, \ldots, \eta_\beta^{\binom{n}{2}}\}$. Assuming that $\eta_\beta(p, q) \leqslant \eta_\beta(u, v)\ \forall\ (p, q) \notin E, (u, v) \in E$, we can approximate $\hat{\tau}(\beta) \approx \eta_\beta^{(m)}$ since we have $m$ edges. Substituting for $\tau(\beta)$ in (3.7), we approximate the gap $\hat{\gamma}(\beta)$. The values of $\hat{\tau}(\beta), \hat{\gamma}(\beta)$ were collected for $\beta \in \{3, 5, 7\}$ and the histogram is shown in Fig. 3.4. Our assumption should be termed as fair, if the gap values lie fairly far away from the origin.

Approximate thresholds show a single prominent peak and light tail, indicating a convergence in the threshold values. And the histograms of gap values (in Fig. 3.4) show that there is a comfortable margin between threshold and expected response. It suggests that, for our distribution from a subset of the *Quick-draw* dataset, there indeed *is a clear separation in the response* *values between the pairs of vertices representing edges and the pairs without edges.*

But, our experiments reveal that the threshold parameter $\tau$ varies with vertex pairs (refer § 3.4.3). To this end, we adapt the threshold to be a per-pair parameter $\tau_{(u,v)}$, and update it using a feedback loop. After every step $i$ of edge inference, we rasterize the geometry $G^{(i)} = (V, E^{(i)})$ using OpenCV drawing functions, as $\hat{I}^{(i)}$, and compute the residual, $\tilde{I}^{(i)} = I - \hat{I}^{(i)}$. A positive value of residual for a pixel $\mathbf{x}$ indicates a false negative, i.e. it was not included in the estimated geometry, whereas a negative value, indicates a false positive, i.e. it is not a part of the true geometry.

We redefine the iterative setup for the feedback loop as in (3.8), with initial conditions defined using a heuristic threshold hyper-parameter as $E_{\beta}^{(0)} = E(\beta, \tau)$ and $\tau_{(u,v)}^{(0)} = \tau \ \forall \ (u,v)$. This edge definition is the same as earlier defined in (3.5,3.6), except that it is defined for the single iteration with ROI-width as the only hyper-parameter; and that threshold is iteratively updated.

At each step after the feedback, the thresholds are updated as in (3.9). This update increments/ decrements the value of the threshold by a fixed amount. A negative residual, implies a false positive, and hence leads to an increment in threshold and vice-versa. The terminal condition is heuristically chosen to be $i_{\max}$ number of iterations.

$$E_{\beta}^{(i)} \equiv \{(u, v) \ : \ \eta_{\beta}(u, v) > \tau_{(u,v)}^{(i)}, u, v \in V, u \neq v\} \tag{3.8}$$

$$\tau_{(u,v)}^{(i+1)} = \tau_{(u,v)}^{(i)} + \lambda \delta^{(i)}(u, v) \tag{3.9}$$

$$\delta^{(i)}(u, v) = \begin{cases} 1, & \text{if} \quad \max(\tilde{I}^{(i)}(u), \tilde{I}^{(i)}(v)) < 0; \\ -1, & \text{if} \quad \min(\tilde{I}^{(i)}(u), \tilde{I}^{(i)}(v)) > 0; \\ 0, & \text{otherwise.} \end{cases}$$

It may be argued that instead of recomputation of edge criterion based on an updated threshold parameter for all distinct vertex pairs, it should be more efficient to simply include the false negatives and exclude the false positives in each update step. As a counter argument, we can see that updating the threshold although expensive, leads to finer steps towards convergence, as against the inclusion-exclusion strategy, which is coarse in nature. The same was confirmed by our experiments, and thus, we chose the threshold update based strategy.

### 3.3.3   Strokes and Gcode

---

**Algorithm 1** `Get_Sequence` Recursive Procedure for generating Sequence of a Stroke

---

1: **procedure** GET_SEQUENCE(A,u,S,strokes_cnt)
    ▷ A is an adjacency list for N vertices.  S is a list of strokes.  strokes_cnt keeps track of total number of strokes.  **pop_edge(u,w)** removes the edge (u,w) from the graph. A[u].**get_vertex()** returns a vertex from adjacency list of u.
2:
3:    If $A[u] ==$ **Null**
4:       **return**
5:    $w \leftarrow A[u]$.**get_vertex()**
6:    **pop_edge(u,w)**
7:    S[strokes_cnt].**append(w)**
8:    GET_SEQUENCE($A, w, S, strokes\_cnt$)
9: **end procedure**

---

---

**Algorithm 2** `Strokes_Gen` Procedure for generating strokes from undirected graph

---

1: **procedure** STROKES_GEN(A)
    ▷ Strokes_Gen procedure called with adjacency list A
2:
3:    strokes_cnt = 0
4:    **while** ($A[v]$ ! **Null** $\forall v$) **do**
5:       S[strokes_cnt].**append(v)**
6:       $u \leftarrow A[v]$.**get_vertex()**
7:       S[strokes_cnt].**append(u)**
8:       **pop_edge(v,u)**
9:       GET_SEQUENCE($A, u, S, strokes\_cnt$)
10:      strokes_cnt $\leftarrow$ strokes_cnt + 1
11:    **end while**
12: **end procedure**

---

In order to compute the strokes, we follow a recursive approach, so that every edge that is traversed is popped out of the graph, and revisiting the vertex is allowed. See Algorithms 1, 2. The time complexity of the algorithm is $\mathcal{O}(|E||V|)$. The strokes are finally translated to GCODE[2] using a simple translation method, *i.e. issue for each stroke in the set, the following commands sequentially*,

- disengage (`G00 Z-5`)

- push first vertex, (`Xff.ff Yff.ff`),

- engage (`G01 Z0`),

- push rest of the vertices (`Xff.ff Yff.ff...`), and

- disengage (`G00 Z-5`)

---

[2] GCODE is a standard numerical control programming language generally used as instructions to a CNC machine. Refer https://en.wikipedia.org/wiki/G-code for more information.

There are two caveats to remember here. Firstly, to disengage (`G00 Z-5`) at the start. And secondly to rescale the coordinates as per machine, for example, `Z-5` pertains to 5 mm away from work, which in case of cm units or inch units, should be scaled by an appropriate factor.

A very similar method is used to render to an SVG Path [Wor18], where each stroke is a series of 2D points, expressed in XML as, `<path d="M x0 y0 L x1 y1 [L x y ...]" />` This is input to the robotic arm to draw on paper.

## 3.4 Experimentation

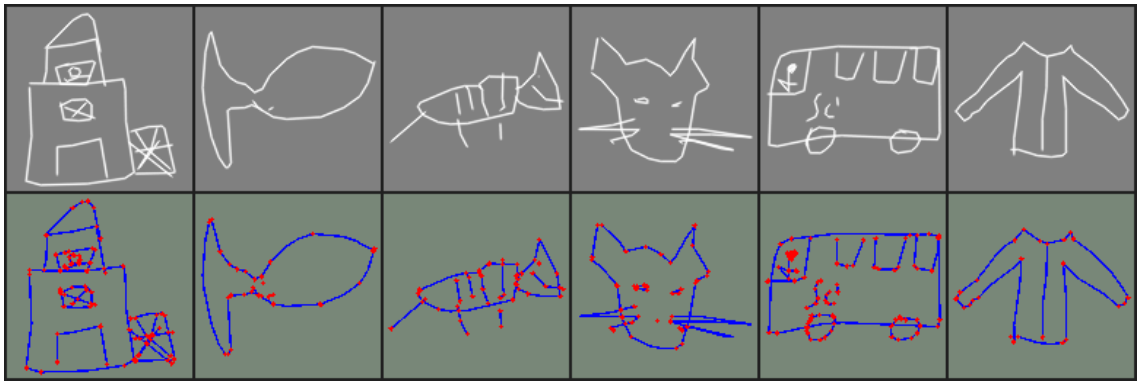

Figure 3.5: Qualitative results of segmentation. *Top.* Samples from the *Quick-draw* dataset; *Bottom.* Rendered true labels.

### 3.4.1 Dataset

We use the *Quick-draw* dataset [HE17] for the basic line art. The sketches are available to us in the form of a list of strokes, and each stroke is a sequence of points. The data is normalised, centred to canvas, and re-scaled to accommodate for stroke width, before being rasterised, using drawing functions, in order to synthesise the input image, $I; I \in \mathbb{R}^{s \times s}$.

For the purpose of segmentation, an active pixel is either identified as a vertex, an edge, or the background; i.e. $K = 3$ classes. We compute the intersections using the vector data from *Quick-draw*, using line sweep algorithm [Ber08] and append them to the accumulated set of vertices, so that they are rasterised as $M_{\text{vertex}}^{\text{GT}}$. The difference operation, with respect to the input, gives us the lines channel, $M_{\text{edge}}^{\text{GT}} = I - M_{\text{vertex}}^{\text{GT}}$. Everything else is the background, $M_{\text{bg}}^{\text{GT}} = 1 - (M_{\text{edge}}^{\text{GT}} + M_{\text{vertex}}^{\text{GT}})$. Thus, we obtain the true labels, $M^{\text{GT}} \in \mathbb{R}^{K \times s \times s}$. A glimpse of the dataset can be seen in Fig. 3.5.

In order to ensure in the trained model, robustness to variations arising out of minor geometric transformations, the dataset had been augmented on the fly, by sampling a random variable for each of: *a)* translation along either axis by a factor of $-\frac{1}{20} < f_t < \frac{1}{20}$ of the canvas size; *b)* a binary horizontal flip; *c)* rotation by $-\frac{1}{20} < \vartheta < \frac{1}{20}$ of circle either ways; and *d)* scaling by a factor of $-\frac{1}{20} < f_s < \frac{1}{20}$ of the canvas size. In implementation, for efficiency of storage, we apply the augmentation on the fly, instead of storing redundant copies of the image.

Concluding, we created a set of $\sim 65k$ unique samples for the training, and $\sim 10k$ samples for testing; each sample being a pair of images, namely the input image and the true labels.

### 3.4.2 Segmentation

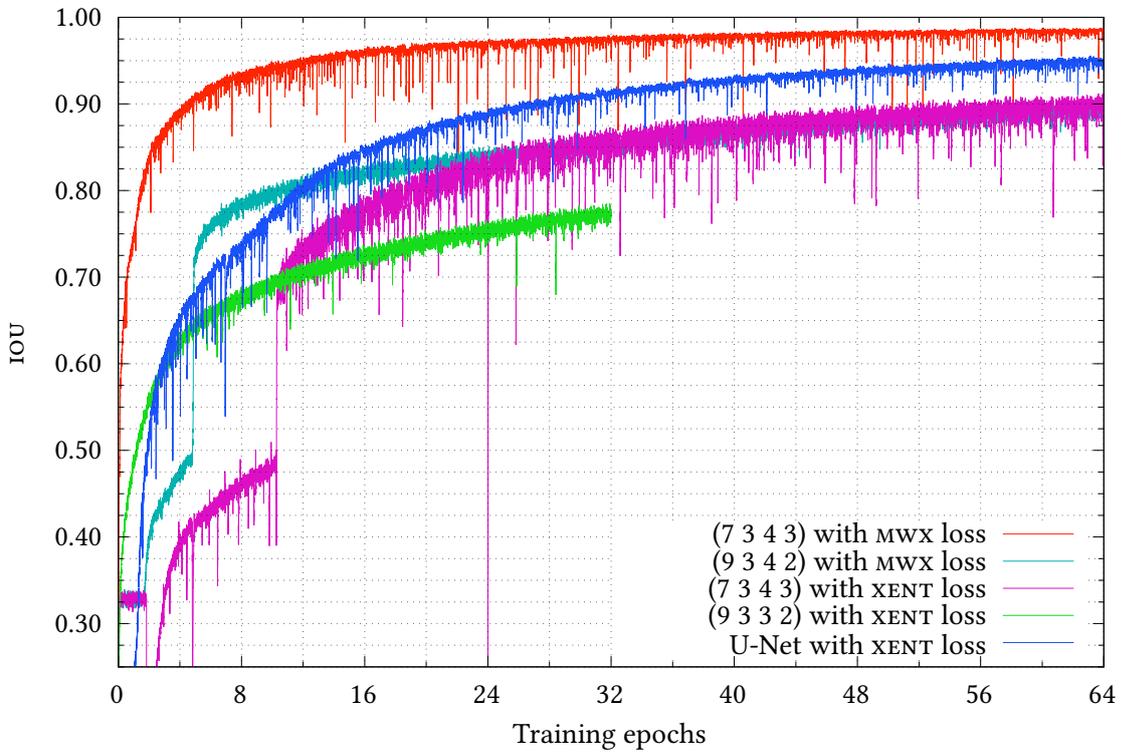

Figure 3.6: Plot of Validation Accuracy (IOU's) using models check-pointed during the training.

Table 3.1: Accuracy measure of segmentation networks

| Architecture | Loss Function | IOU (%) |
|---|---|---|
| Net (7 3 4 3) | MWX | **98.02%** |
| Net (9 3 4 2) | MWX | 91.24% |
| Net (7 3 4 3) | XENT | 92.57% |
| Net (9 3 3 2) | XENT | 75.09% |
| Vanilla U-Net | XENT | 94.23% |

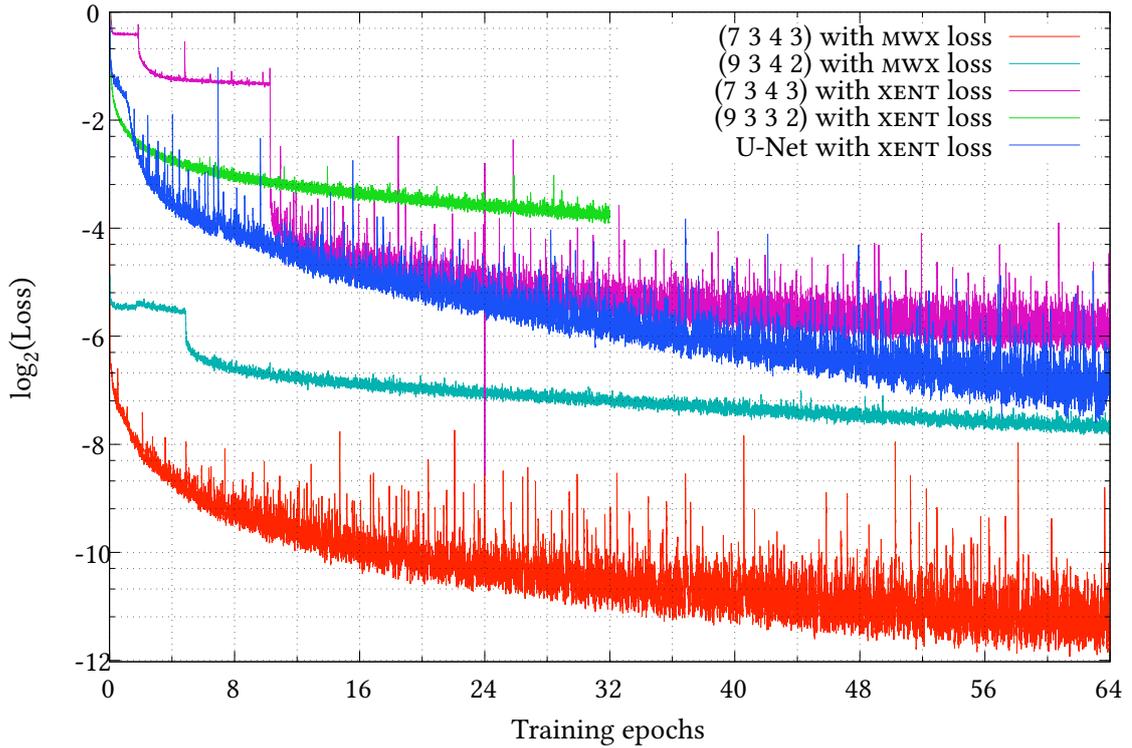

Figure 3.7: Plot of Validation Loss in logscale using models check-pointed during the training.

We experimented with various combinations of network architectures, and training objectives, to find a suitable segmentation network. The most promising ones that have been presented here are: *a)* Net (9 3 2 2) with cross-entropy loss (xent); *b)* Net (9 3 3 2) with xent; *c)* Net (7 3 4 3) with xent; *d)* Net (7 3 4 3) with max-weighted cross-entropy loss (mwx). (For network nomenclature, refer § 3.3.1 and Fig. 3.2.) Additionally, we also trained a variant of vanilla U-Net with xent, where we replaced 'valid' convolutions with padded-convolutions to obtain consistent image sizes on our dataset. As per our nomenclature, the vanilla U-Net corresponds to (3 3 4 2) architecture. The optimiser used was RMSProp [HSS], with $\beta = 0.9$.

The plot of validation accuracy and losses during the training are reported in the Fig. 3.6, 3.7. We can see that net (7 3 4 3) with mwx stabilises fastest and promises the most accurate results. A quantitative summary of our evaluation over the test set, is shown in the Table 3.1. Here also we see a similar trend, *i.e.* (7 3 4 3) with mwx outperforms others by far. Qualitatively, the implication of a better segmentation is most prominently visible in fairly complicated cases, *e.g.* Fig. 3.3. This improvement in the results is attributed to the design of max-weighted cross entropy loss (refer § 3.3.1).

### 3.4.3   Graph Inference

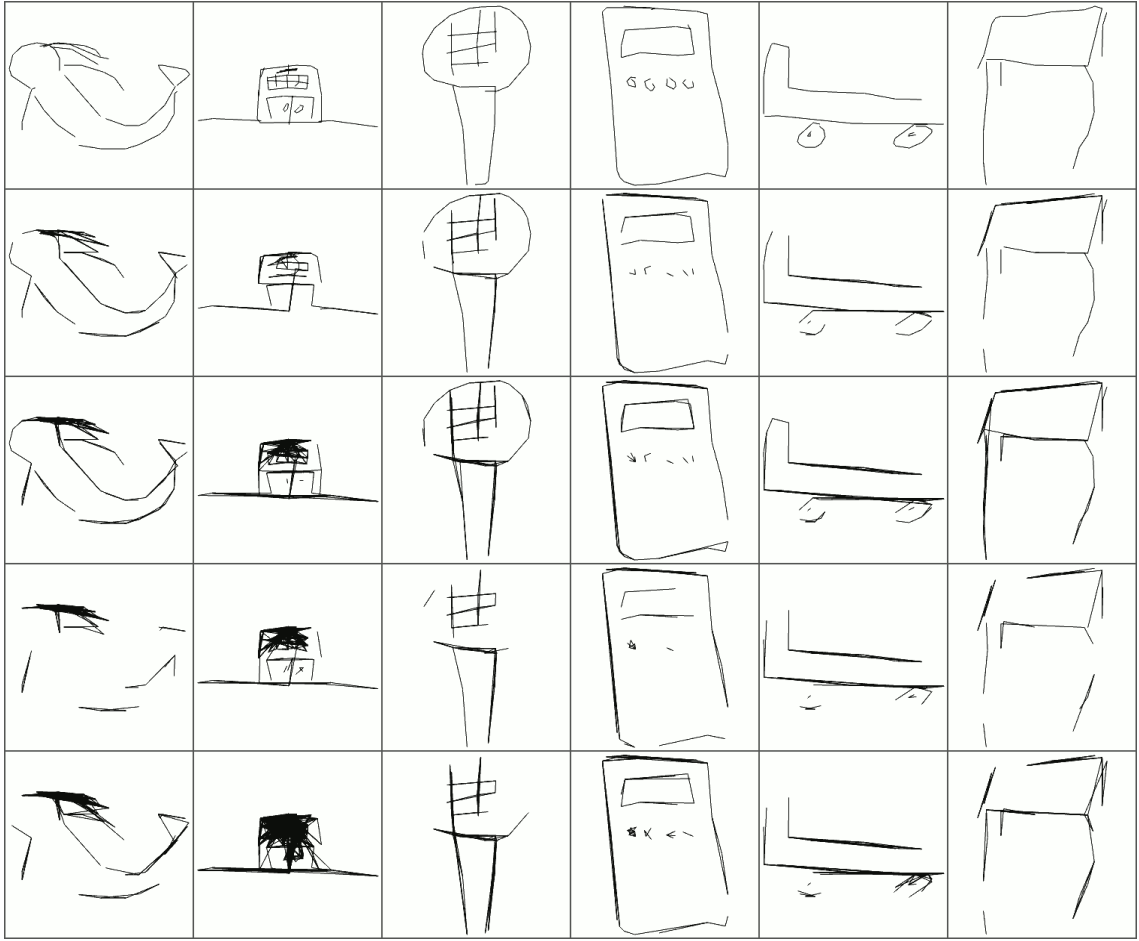

Figure 3.8: Qualitative evaluation with naïve edge inference method as in (3.4, 3.5, 3.6). The top row shows the input. The subsequent rows show reconstruction using a graph interpreted from segmentation mask, with values of mask width and threshold $(\beta, \tau)$ set as (2, 0.3), (3, 0.22), (5, 0.2), (7, 0.15) respectively downwards.

We qualitatively evaluated the method of naïve edge inference in (3.4, 3.5, 3.6) with values of mask-width and a fixed plausibility threshold, $(\beta, \tau) \in \{(2, 0.3), (3, 0.22), (5, 0.2), (7, 0.15)\}$. The results are shown in Fig. 3.8. We see that a fixed value of threshold is not able to capture the edges based on the plausibility score.

Thereafter, we tested the adaptive update method, designed in (3.8, 3.9), with heuristically set parameters: mask width $\beta = 1.8$, initial plausibility threshold $\tau = 0.35$, update rate $\lambda = 0.05$. Typically, we found satisfactory results with $i_{\max} = 10$ updates. In Fig. 3.9, the second row is a rendering of naïve graph inference after $i = 0$ updates, and has missed a few strokes. In almost all the cases after $i = i_{\max} = 10$ updates (bottom row) the missing strokes have been fully recovered, and extra strokes have been removed.

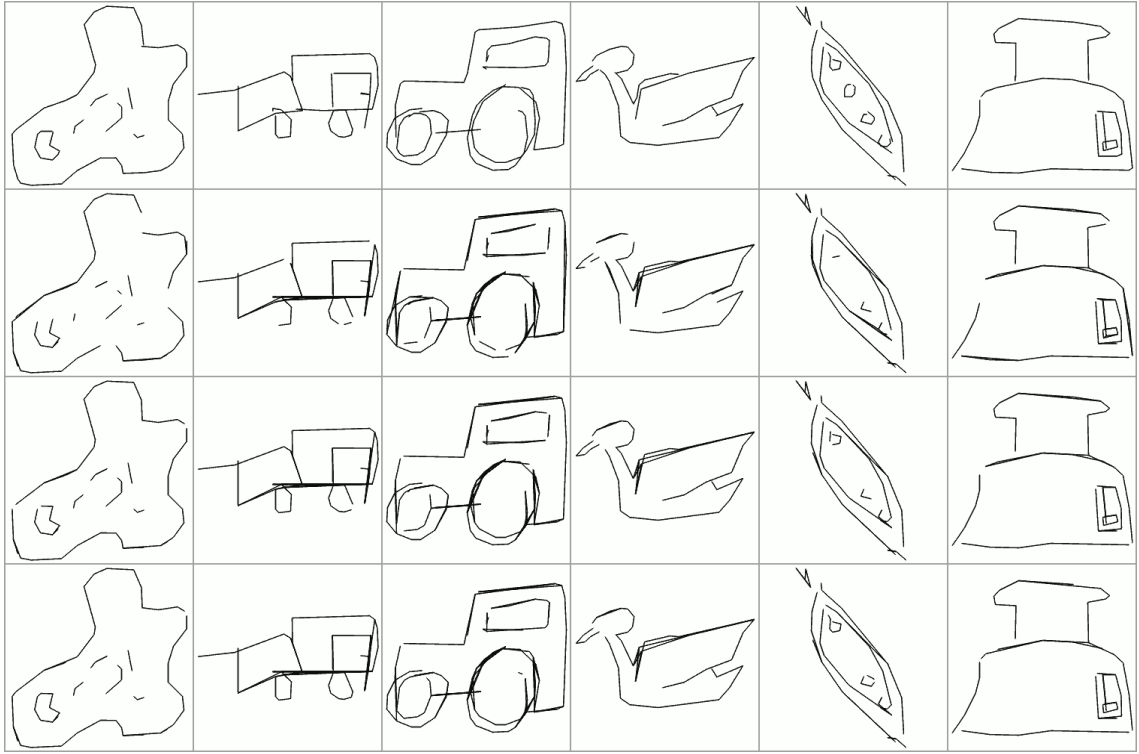

Figure 3.9: The results of feedback loop in graph interpretation, using (3.8, 3.9). *Rows top to bottom.* the input image; naïve graph interpretation: $i = 0$ updates; interpretation after $i = 5$ updates; and after $i = 10$ updates.

We also see some limitations in a graph interpretation from image domain; for example in Fig. 3.9, even after $i = 10$ updates, there are some overlapping lines that haven't been removed. The point is clearer in the video on the project page[3], where the pen repeats the same line.

### 3.4.4  Application

We tested the results for application with two robots. First is a traditional CNC milling machine "EMCO Concept Mill 250," where we used pen as a tool to plot over a plane surface. This machine accepts GCODE as input. Second is a modern robotic arm "Dobot Magician Robotic Arm," that holds a pen to draw on paper. This machine accepts line drawings in SVG format, such that each stroke is defined as a `<path>` element. The method of post processing the list of strokes to either create GCODE for CNC plotter or create SVG for a robotic arm is detailed out in § 3.3.3. For practical purposes, we fixed the drawings in either case to be bounded within a box of size 64mm × 64mm and set the origin at (25mm, 25mm) Fig. 3.11 shows a still photograph of the robots in action. Reader is encouraged to watch the video on the project page[4] for a better demonstration. Both the robots, namely the CNC plotter and the robotic arm successfully complete the task assigned to them, indicating that the stroke extraction and their translation was correct; and also that the complete pipeline works correctly as expected.

---

[3]https://bvraghav.com/can-i-teach-a-robot-to-replicate-a-line-art/
[4]https://bvraghav.com/can-i-teach-a-robot-to-replicate-a-line-art/

### 3.4.5 Generalisation

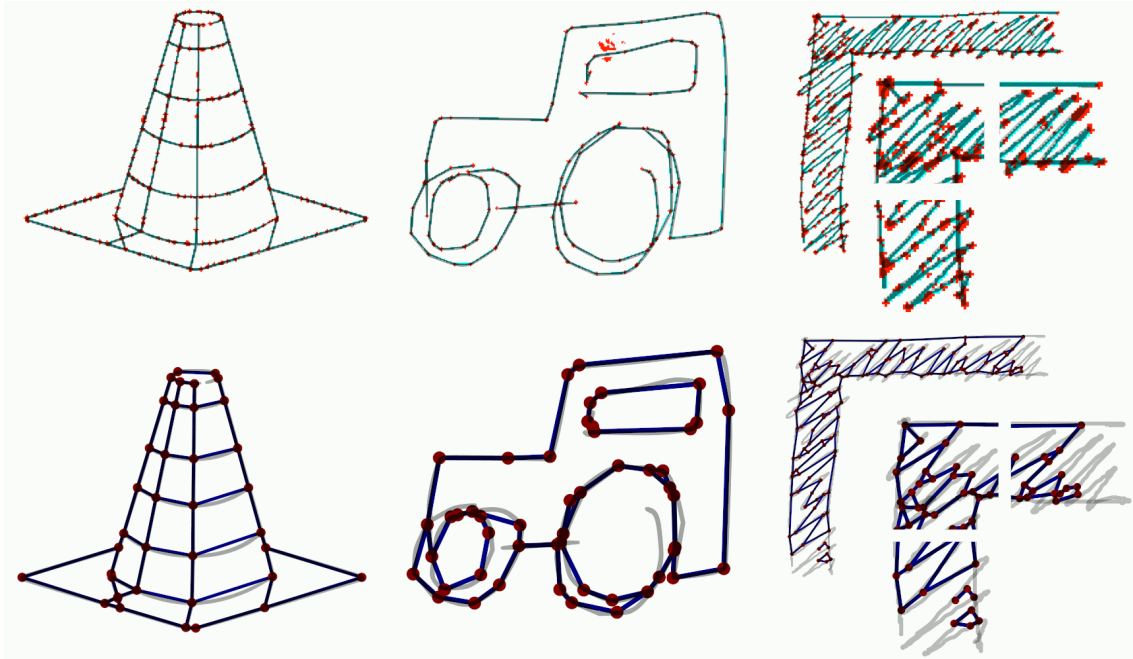

Figure 3.10: Comparison of our segmentation (left column) against Favreau's vectorization (right column). *Top to Bottom.* A simple line sketch from Favreau's examples; A simple line sketch from *Quick-draw* dataset; A complicated hatch pattern with zoomed in details.

A deep learning based method is generally preferred for it generalises well. Here we qualitatively inspect our model against Favreau *et al.*'s vectorization method [FLB16] to line drawing vectorization, using three different samples, namely a simple sketch from their samples; a simple sketch from *Quick-draw*; and a complicated sketch, representing a hatch pattern, commonly used to represent walls in architectural graphics. In Fig. 3.10, we see that our method produces satisfactory segmentation on both the simple cases, and also gracefully degrades its performance in the complicated case; whereas, the latter method, compromises on details, *e.g.* few open strokes near the wheels of the automobile, and thus fails miserably on the complicated task of vectorizing a hatch pattern, missing out on several details.

## 3.5 Discussion

This chapter has focused on one of the problems in structural interpretation of sketches in 2D, and broken it into three stages as,

$$\text{Sketch} \rightarrow \{\text{edge mask + vertex mask}\} \rightarrow G(V, E) \rightarrow \text{Strokes}.$$

Having distilled an explicit graph structure $G(V, E)$ in the pipeline has the advantage of leveraging over a body of theoretically grounded mathematical work for downstream task. Say for example, *deciding upon the similarity* between two sketches on the basis of their respectively distilled graph structure representations. Our method to the geometric transformations is robust, and thus provides some degree of reliability within a given domain. But as common in deep-learning based methods, we see degradation when used along with out-of-domain images.

**Sketch to geometry correspondence.** In general Sketch $\rightarrow G(V, E)$ is an ill posed problem. Because for a given graph $G$, there are many plausible sketches that correspond to it, depending upon the style of rasterisation; and for a given sketch theorised as an ideal geometry, there are many plausible graphs that correspond to it depending upon the hyperparameters of approximation algorithm used in order to create a piece-wise linear geometry. Hence the problem of sketch $\leftrightarrow$ geometry correspondence is open to interpretation, and generally resolved using domain-specific heuristics or through the end user control. In some sense our trained model represents a specific hyperparameter setting of the sketch $\rightarrow$ geometry $\rightarrow$ graph approximator, where the geometry is latent and by observing the graph output, we assert its comparability to the state-of-the-art.

**Replaceable modules in the pipeline.** To take it further, the three-stage pipeline is open to investigation using state-of-the-art methods. For example, using conditional random field (CRF) based methods for segmentation [ST16]; or using deep-learning (DL) in conjunction with CRF for segmentation; or using graphical models for graph interpretation from segmentation masks; or seeking a further more efficient $\mathcal{O}(|V| + |E|)$ algorithm for inferring strokes. This chapter thus opens up a different avenue for investigation in skeletonisation and vectorisation of sketches.

**End to end inference.** In the graph interpretation step, the edge detection, which has been formulated as a decision problem, may as well be formulated as a classification problem, taking the graph structure estimation one step closer to being trained from end to end. Also formulating the problem on the lines of multiple object detection should be worth investigating and forms a future scope of this thesis.

## 3.6   Conclusion

We have portrayed an alternate method for a drawing robot in the context of line drawings, which maps image domain to a graph structure that represents the original image, and retrieve a set of actionable sequences.

There is definitely, a scope of improvement here. We see the following points to be most significant,

- The segmentation model is highly sensitive to noise, and similar phenomenon has been observed in the context of deep learning by Goodfellow et.al [GSS14]. This problem is subject of contemporary research.

- Graph interpretation method is sensitive to initial choice of threshold parameter $\tau$ and the update hyperparameter $\lambda$, refer (3.9).

The pipeline proposed by us is a multi utility technique, the use of which we have portrayed through a CNC machine as well as through a robotic arm as shown in Fig. 3.11

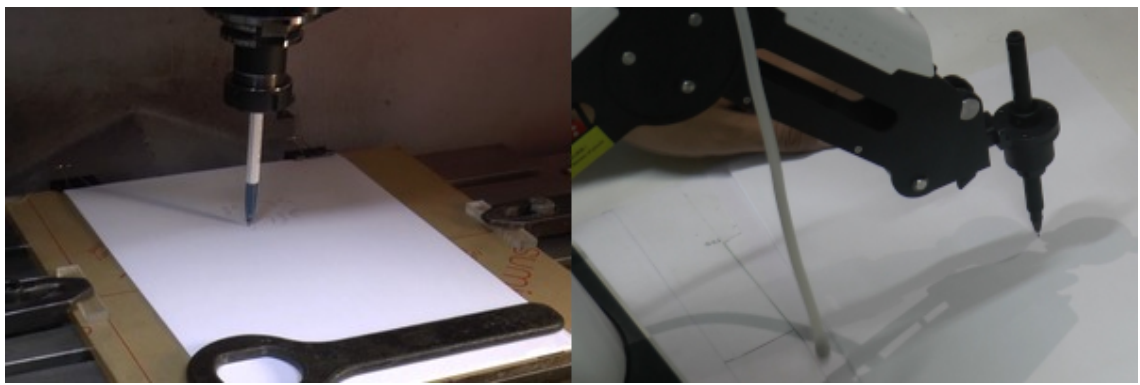

Figure 3.11: Demonstration of the applicability of our 2D prediction output using two different machines, namely, the CNC plotter in the left, and the robotic arm in the right.

CHAPTER# 4

# 3D Prediction

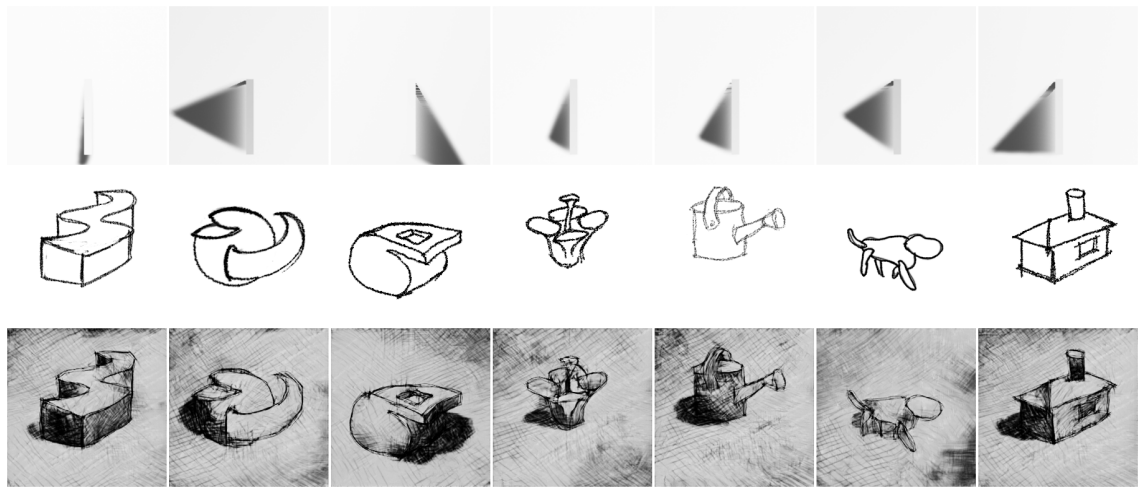

Figure 4.1: SHAD3S: a Model for Sketch, Shade and Shadow — is a completion framework that provides realistic hatching for an input line drawn sketch that is consistent with the underlying 3D and a user specified illumination. The figure above shows choice of illumination conditions on the top row, user sketches in the middle row, and the completion suggestions by our system. Note that the wide variety of shapes drawn by the user, as well as the wide variations in brush styles used by the user were not available in the training set.

In this chapter, we test the 3D-geometric understanding of a sketch-based system without explicit access to the information about 3D-geometry. The objective is to complete a contour-like sketch of a 3D-object, with illumination and texture information as if it a human would accentuate its 3D nature using artistic hatch patterns. We propose a data-driven approach to learn a conditional distribution modelled as deep convolutional neural networks to be trained under an adversarial setting; and we validate it against a human-in-the-loop. The method itself is further supported by synthetic data generation using constructive solid geometry following a standard graphics pipeline. In order to validate the efficacy of our method, we design a user-interface plugged into a popular sketch-based workflow, and setup a simple task-based exercise, for an artist. Thereafter, we also discover that form-exploration is an additional utility of our application.

Sketches in some sense form a lower bound of graphical human expression. And expression of a 3D spatial setting is essential to multiple disciplines, *e.g.* Design, Architecture, Engineering, that support various industries as the likes of construction and manufacturing. Hatching is a common method used by artists to accentuate the third dimension of a sketch, and to illuminate the scene.

Our system SHAD3S[1] attempts to compete with a human at expressing generic three dimensional (3D) shapes on a two-dimensional (2D) canvas. Not surprisingly, this is of assistance to an artist in a form exploration exercise. The novelty of our approach lies in the fact that we make no assumptions about the input sketch on a 2D canvas, other than that it represents a 3D shape, and yet, given a contextual information of illumination setting and texture filling as a hint, our model is able to synthesise an accurate hatch pattern over the sketch, without access to 3D or pseudo 3D. In the process, we contribute towards *a)* a cheap yet effective method to synthesise a sufficiently large high fidelity dataset, pertinent to task; *b)* creating a pipeline with conditional generative adversarial network (CGAN); and *c)* creating an interactive utility with GIMP, that is a tool for artists to engage with automated hatching or a form-exploration exercise. User evaluation of the system suggests that the model performance does generalise satisfactorily over diverse input, both in terms of style as well as shape. A simple comparison of inception scores (§ 4.7.2) suggest that diversity in the generated distribution satisfactorily matches that of the ground truth.

---

[1] `https://bvraghav.com/shad3s/`— The project page; hosted with further resources.

## 4.1   Introduction

Sketches are a widely used representation for designing new objects, visualizing a 3D scene, for entertainment and various other use-cases. There have been a number of tools that have been developed to ease the workflow for the artists. These include impressive work in terms of non-photorealistic rendering (NPR) that obtain sketches given 3D models [HZ00; GI13], others that solve for obtaining 3D models given sketch as an input using interaction [HGY17; DAI+18], and several others that aim to ease the animation task [BCK+13; JFA+15; FJS+17]. However, these works still rely on the input sketch being fully generated by the artist. One of the important requirements to obtain a good realistic sketch is the need to provide the hatching that conveys the 3D and illumination information. This is a time consuming task and requires effort from the artist to generate the hatching that is consistent with the 3D shape and lighting for each sketch drawn. Towards easing this task, we propose a method that provides realistic hatching for an input line drawn sketch that is consistent with the underlying 3D and illumination.

Though recent works have tried to address problems associated with shading sketch [ZLB20], however, to the best of our study we couldn't find any prior work addressing the problem we intend to solve. Here we hope to leverage deep learning to decode and translate the underlying 3D information amongst a rasterised set of 2D lines. Deep learning has been promising to solve many interesting and challenging problems [GPM+14; MO14; IZZE16; DAI+18]. Moreover, it is well known that deep learning algorithms are data intensive. There exist datasets [DAI+18; ZLB20] which aim to solve 3D-related problems like predicting the illumination, and mesh inference from line drawings, but they are limited in their diversity and level of pertinence (see § 4.2.3). This motivates us to contribute the SHAD3s dataset for *Sketch-Shade-Shadow* like tasks (see § 4.4). Hereby, we attempt to help the research community to bring closer, the two domains of graphics and vision.

**An overview of the problem statement.**   With the aim of creating artistic shadows with hatch-patterns for a given contour drawing of a 3D object, under a user specified illumination, we define this research problem as an attempt *to learn a function*, that maps a combination of — *a)* hand-drawn 3D objects; *b)* illumination conditions; and *c)* textures — to a space of completed sketches with shadows manifested as hatch-patterns. See § 4.3.

**An overview of the proposed solution.**   Our proposed pipeline is based on training a conditional generative adversarial network (CGAN) [MO14] with a data-set of 3D shape primitives and aims to model the problem as one of contextual scene completion. While this approach has been widely explored in the image domain [LSD15; IZZE16], the challenge is to ensure that we are able to obtain convincing results in sketch domain using a sparse set of lines. In order to solve this we train our model using a novel SHAD3S dataset explained further in § 4.4. These are rendered to be consistent with the illumination and 3D shape. We include the context required for solving the problem in the input. Once this is learned it is possible to automate the sketch generation using the proposed CGAN model. A glimpse of our results can be seen in Fig. 4.1, that once again reinforces the significance of sufficiently large dataset in the context of deep learning, and its generalizability.

A natural approach given a dataset would be to train a regression based system that would aim to generate the exact information that is missing. This can be achieved using reconstruction based losses. The drawback of such an approach is that the resultant generation would be blurred as it averages over the multiple outputs that could be generated. In contrast, our adversarial learning based approach allows us to generate a sharp realistic generation that we show is remarkably close to the actual hatch lines.

The system has been tested for usability and deployment under casual as well as involved circumstances among more than 50 participants, the qualitative results of which are selectively displayed here (see § 4.8), and a larger subset has been made available as supplement. We would like to mention that the samples for which the system has been tested by the user *are very different* from the distribution of samples for which it was trained on. It is able to generalize and produce satisfactory sketches as shown in Fig. 4.1. Through this paper we provide, to the best of our knowledge, the first such tool that can be used by artists to automate the tedious task of hatching each sketch input. The tool and the synthetic dataset (described in § 4.4) are available for public use at the project page[2].

**Summary.**   To provide an overview of our work, we summarise our main contributions as follows. *Firstly,* we define an extremely inexpensive method to generate sufficiently large number of high fidelity stroke data, and contribute the resultant novel public dataset.

*Secondly,* we define simple method to inject data from three different modalities, namely: *i)*

---

[2] https://bvraghav.com/shad3s/

sketch representing 3D information; *ii)* rendered canonical geometry representing illumination; and *iii)* a set of hand drawn textures representing multiple levels of illumination — these three may be plugged into any well established deep generative model.

*Finally,* we push boundaries for artists through an interactive tool to automate a tedious hatching task, and to serve as a form-exploration assistant.

## 4.2   Relevant Works

As the proposed method is the first work to address the task of automating hatch generation from sketches, there is no directly related work to the best of our knowledge. However, there are two prominent lines of approach, namely stylization and sketch-based modeling that are related to our problem. In contrast to these approaches, we aim to address the task as one of contextual information completion and therefore the proposed approach differs substantially from either of these approaches.

### 4.2.1   Stylization

**In the early works,**   an undercurrent is arguably visible in a series of works following Siggraph '88, where in a panel proceedings [MFB+88], Hagen said,

> The goal of effective representational image making, whether you paint in oil or in numbers, is to select and manipulate visual information in order to direct the viewer's attention and determine the viewer's perception.

We see this in the initial works [Hae90; SABS94; CAS+97] focusing on attribute sampling from user input to recreate the image with arguably a different visual language and accentuation and offer interactivity as a medium of control. Thereby followed, a shift towards higher order controls, to achieve finer details, for example temporal coherence [Lit97], modelling brush strokes [Her98], and space-filling [SY00]. Space-filling had been buttressed, on one hand with visual-fixation data and visual-acuity [DS02; SD02], and on the other with semantic units inferred using classical vision techniques [ZZXZ09]. The finer levels of control, in these works, allowed for interactivity with detail in different parts of the synthesised image.

**Image Analogies**   had popularised the machine learning framework for stylising images, with analogous reasoning $A : A' :: B : B'$, to synthesise $B'$, using image matching [HJO+01];

to achieve temporal coherence [BCK+13]; using synthesis over retrieved best-matching feature [FJS+17; FJL+16; JFA+15]. The problems thus formulated, allowed use of high level information as a reference to a style, for example a Picasso's painting.

**Deep Learning Techniques** have recently documented success in similar context, using Conditional Generative Adversarial Networks (CGAN) [MO14], and its variants using sparse information [IZZE16]; using categorical data [WLZ+17]; or for *cartoonization* of photographs [CLL18].

Researchers have recently contributed to line drawings problem pertaining to sketch simplification [SISI16; SII18a], line drawing colorization [FHOO17; KJPY19; ZLW+18] and line stylization [LFH+19]. Sketch simplification [SISI16; SII18a] aims to clean up rough sketches by removing redundant lines and connecting irregular lines. Researchers take a step ahead to develop a tool [SII18b] to improve upon sketch simplification in real-time by incorporating user input. It facilitates the users to draw strokes indicating where they want to add or remove lines, and then the model will output a simplified sketch. *Tag2Pix* [KJPY19] aims to use GANs based architecture to colorize line drawing. *Im2pencil* [LFH+19] introduce a two-stream deep learning model to transform the line drawings to pencil drawings. Among the relevant literature we studied, a recent work [ZLB20] seems to be most related to our approach, where authors propose a method to generate detailed and accurate artistic shadows from pairs of line drawing sketches and lighting directions.

In general, it has been argued that models over the deep learning paradigm handle more complex variations, but they also require large training data.

### 4.2.2 Sketch-based Modeling

**Interactive Modeling** [ZHH96] introduced a set of rule-based shortcuts with the aim of creating *an environment for rapidly conceptualizing and editing approximate 3D scenes.* Sketching was integrated soon into the system with TEDDY and its relatives, [IMT99; IH03; MI07], with prime focus on inflated smooth shapes as first class citizens, and sketch-based gestures for interactivity. Recently, SMARTCANVAS [ZLDM16] extended the works relying on planar strokes.

**Analytical Inference,** for 3D reconstruction is another popular approach to modeling from sketches. With the context of vector line art, [MM89] inferred simple curved shapes, and [LS96] showed progress with complex compositions of planar surfaces. Recently, for illuminating sketches, [SBSS12; IBB15] had shown the effectiveness of normal-field inference through regularity cues, while [XGS15] used isophotes to infer the normal-field.

**Deep Learning Techniques,** have more recently, shown substantial progress in inferring 3D models from sketches: a deep regression network was used to infer the parameters of a bilinear morph over a face mesh [HGY17] training over the *FaceWarehouse* dataset [CWZ+14]; a generative encoder-decoder framework was used to infer a fixed-size point-cloud [FSG17] from a real world photograph using a distance metric, trained over *ShapeNet* dataset [CFG+15]; U-Net [RFB15]-like networks were proven effective, to predict a volumetric model from a single and multiple views [DAI+18] using a diverse dataset, but this dataset *lacks pertinence to our task.*

### 4.2.3 Our contribution

**Dataset.** Deep learning applications have stepped into many research domains and the profusion of data has nicely complemented deep learning algorithms. However, there does seem to be lack of feasible, scalable, impressive in terms of both quality and quantity, and user-interpretable dataset in the graphics community specially artistic domain. There does seem to be a gap when we compare these datasets with popular datasets suitable for deep learning algorithms. For instance, ShadeSketch dataset [ZLB20] *doesn't offer much diversity* as it is limited to the domain of manga art and dataset has only 1160 images which is a relatively small number. We thus contribute a dataset (see § 4.4) creation framework with a codebase released into public domain, with an aim towards generalizability of our approach.

**Approach.** The stylization based approaches incorporate image translation without any explicit access to information of geometry, but require dense input. The geometric modeling based approaches use sparse input but are not robust enough for downstream graphic processing. We take the best of two worlds, to formulate the problem of generating hatches using our contextual sketch completion framework, in our approach. The framework ensures that the relevant context required in terms of illumination information and textures is incorporated. We use generative adversarial networks to ensure realistic generation of hatch patterns that are indistinguishable from ground-truth hatch patterns that are generated by using accurate ground-truth 3D. Note that in our inference procedure, we make use only of the line drawings, the illumination and the texture context. We *do not* require access to 3D or psuedo-3D in order to generate the accurate hatch pattern.

**Framework.** Our framework stands out in the following ways.

*Firstly*, the problem of stylization inherently requires an information-dense image to start with. For an artist, it is evidently quicker as well as more intuitive, to create a basic line-art, than to create a well rendered sketch. We use the basic line art as input, which is sparse in nature.

*Secondly*, the heuristic based methods although coarsely capture the distribution, wherein different parameter values cater to different class of problems, they are vulnerable in fringe cases. We on the other hand, use deep learning based models, that have shown promise towards robustness and generalization.

*Finally,* our model offers controllable stylization parameters of lighting and texture, that are pivotal for designers in decision making.

## 4.3 Problem formulation

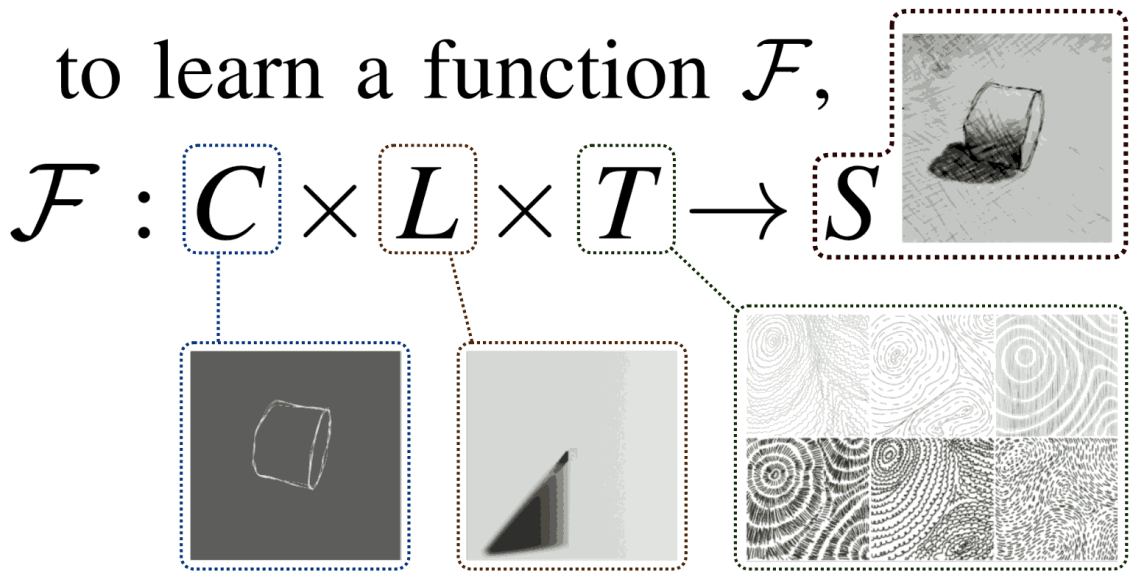

Figure 4.2: The problem formulated as an objective to learn a function $\mathcal{F} : C \times L \times T \rightarrow S$

As illustrated in Fig. 4.3, the objective of this research is to learn a function $\mathcal{F} : C \times L \times T \rightarrow S$; where, $C$ represents the space of hand-drawn contours for any 3D geometry; $L$ represents an illumination analogy shown for a canonical geometry; $T$ is a set of textures representing different shades; and $S$ is the space of completed sketches with shadows manifested as hatch-patterns.

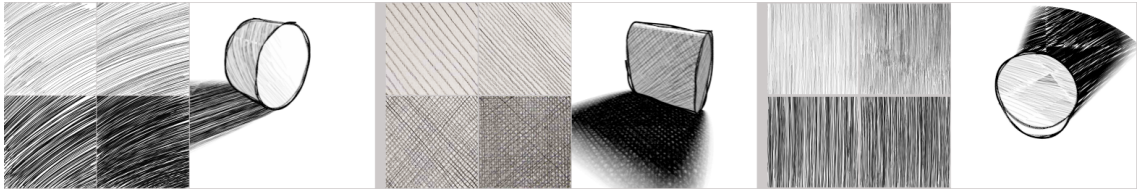

Figure 4.3: Three sketches rendered using three different objects with TAM's implemented using BLENDER.

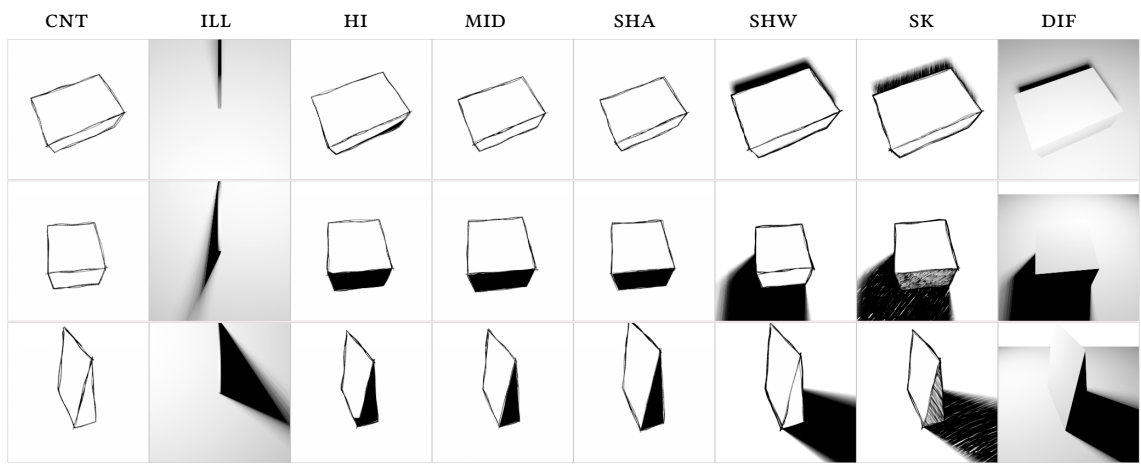

Figure 4.4: Few examples from the dataset. *Columns left to right.* CNT: Contour; ILL: Illumination; HI: Highlight mask; MID: Midtone mask; SHA: Shade mask (on object); SHW: Shadow mask; SK: Sketch render; DIF: Diffuse render.

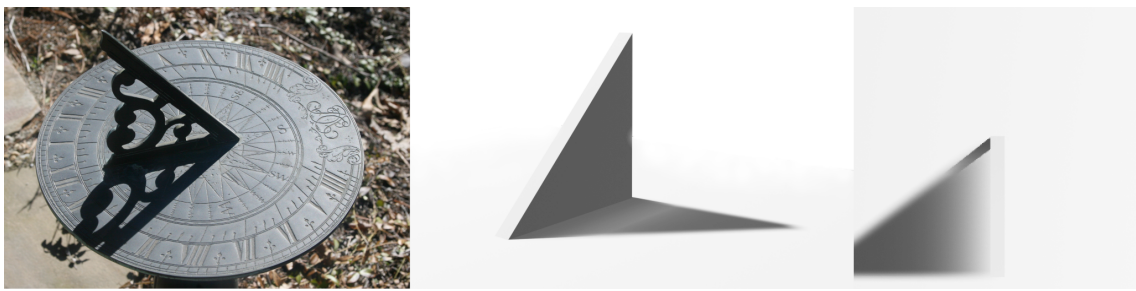

Figure 4.5: Gnomon in a sundial is used as canonical object to capture illumination information. *Left to right:* A sundial (*Courtesy: liz west* https://flic.kr/p/EWBd4); Perspective render of a gnomon; A top view of the same.

## 4.4  Dataset

We introduce the SHAD3S dataset ( Fig. 4.4), where each data-point contains: *a)* contour drawing of a 3D object, corresponding to $\mathbf{c} \in C$; *b)* an illumination analogy; *c)* 3 illumination masks over *(a)*, namely *highlights, midtones* and *shades*; *d)* a shadow mask on the ground; *e)* a diffuse render; and *f)* a sketch render. Additionally, it contains a catalogue of 6 textures.

This is a fully synthetic dataset created from scratch using the following steps: *a)* Create a Constructive Solid Geometry (CSG)-graph using simple grammar over a finite set of eulidean solids; *b)* Create a mesh from CSG; *c)* Render the mesh. The illumination masks were created by thresholding a diffuse shader. Blender's [Ble19b] official freestyle tool [Ble19a] was used to render the contours. And sketch renders were created after implementing the tonal art maps [PHWF01] (see Fig. 4.4).

Since the geometry is also procedurally generated on the fly, we take the opportunity to progressively increase the complexity of geometry by varying the upper-bound of number of solids in the composition from 1 to 6. This assists in the progressive evaluation of the models.

Additionally, we also stored the information of illumination conditions in the form of a diffuse render of a canonical scene with a canonical wedge-like object resembling the gnomon of a sundial. The rationale behind using this style of information is to utilize the image domain.

We also attempted to use a background filled with hatch patterns, instead of white background, and analysed its effects on the model performance detailed out in § 4.7.

For the purpose of this research, we follow the principle of tonal art maps [PHWF01], albeit in a low-fidelity version (see § 4.4.3). To approximate the effect, we create an 6-tone cel-shading [KKY01], including the pure white and pure black. For the rest of the four shades we map one texture corresponding to each tone on the object. The similarity to tonal art maps lies in the fact that any hatch pattern map for a given tone should also be visible in all the tones darker than itself. To this effect, we collect samples of high resolution tonal art maps apriori, from artists. A random crop of these images is used by the model as input conditions (see Fig. 4.4)

All in all, we create *six subsets*, each restricting the maximum number of solids in a CSG composition of a scene to be from *1 through 6.* Each subset is created from $\sim 1024$ distinct scenes. Each scene is rendered from $\sim 64$ different camera poses. The dataset thus contains $\sim 2^{16} \times 6$ data points with a resolution of $\sim 256 \times 256$. **The data synthesis code is publicly available** at the project page: https://bvraghav.com/shad3s/.

### 4.4.1   Solid Grammar

Constructive solid geometry [HDM+13] (CSG) has been a commonly used approach for procedural modeling [EMP+03]. At the core of CSG lies a basic constructivist approach [PI69], so that all geometry may be interpreted as a certain combination of simple primitive geometry. For example, the famous apple logo may be seen as a circular part *subtracted from* a corner of a simple apple figure; the apple figure itself may further be seen as *a union of* two symmetric apple halves and a leaf graphic; the leaf graphic as *a rotated version of* an *intersection of* two circles; and so forth. The description naturally invokes a tree structure in the mind of a computer scientist.

Formerly, CSG was known as computation binary solid geometry for it may be expressed as a binary tree with each leaf node representing a primitive, the edges representing geometric transformations and the other nodes representing binary operations. This may be encoded as a BNF as follows.

$$
\begin{aligned}
\langle\text{Geometry}\rangle \ &\vDash\ \langle\text{Operator}\rangle\langle\text{Geometry}\rangle\langle\text{Geometry}\rangle \\
&\mid\ \langle\text{Transformation}\rangle\langle\text{Geometry}\rangle \\
&\mid\ \langle\text{Primitive}\rangle \\
\langle\text{Operator}\rangle \ &\vDash\ \cup \mid \cap \mid - \\
\langle\text{Primitive}\rangle \ &\vDash\ \textit{a predefined set of geometric primitives} \\
\langle\text{Transformation}\rangle \ &\vDash\ \textit{a predefined set of geometric transformations}
\end{aligned}
$$

In order to generate a complex geometry randomly, from here, we need to specify four control variables, namely *a)* $N$ : number of leaves; *b)* $\{G_i \ : \ 0 < i \leqslant N\}$ transformed geometric primitives, one for each leaf; *c)* $\{R_i \ : \ 0 < i < N\}$ operators, one for each of $N - 1$ nodes. It is noteworthy that, we distilled all the geometric transform from the edges down to the leaf nodes, so that instead of *primitives*, the leaf nodes now store TRANSFORMED PRIMITIVES. This begets a special case when the leaves are transformed primitives, whereas, all the edges, now represent *identity* as transformations. When compared against the general case, the special case has some loss of *expressiveness* of its grammar in representation, but no loss of generality.

Another key aspect of real world data that needs to be accommodated in this procedural modeling is that of *symmetry*. This has earlier also been seen in the models generated for volu-metric prediction [DAI+18]. To this end, we introduce another production rule in our grammar. The two changes to the general grammar may be documented as follows.

$$
\begin{aligned}
\langle \text{Geometry} \rangle \quad &\vDash \quad \langle \text{Operator} \rangle \langle \text{Geometry} \rangle \langle \text{Geometry} \rangle \\
&\mid \quad \langle \text{TransformedPrimitive} \rangle \\
\langle \text{Operator} \rangle \quad &\vDash \quad \cup \mid - \\
\langle \text{TransformedPrimitive} \rangle \quad &\vDash \quad \textit{a set of transformed primitives} \\
&\mid \quad \textit{a set of transformed symmetric primitives}
\end{aligned}
$$

It is noteworthy that, *in order to approximate the real world objects,* we reduce the set of operators to a binary set of either union ($\cup$) or difference ($-$). This is done heuristically, because in our experiments, we found that intersections, more often than not, led to a resulting geometry with sizes of a very small fraction of the operand geometries.

The sampling or each complex, finally, is done by choosing the total count of leaves $N$, from a uniform distribution; followed by choosing the count of symmetric objects from a bernoulli distribution with success rate $f$, so that their expected count, $\mathbb{E}[N_2] = 0.5Nf$ and $N_1 + 2N_2 = N$, where $N_1, N_2$ represent single and symmetric primitives respectively.

We use prefix notation to represent the tree, and indices to represent transformed primitives. So the symmetric primitives are represented as a unit depth subtree as $\cup$ *i j*. The information that *i*-th and *j*-th primitives are symmetric is distilled into the description of primitives, as shown in § 4.4.2. Starting with $N_1 + N_2$ single or symmetric primitives, we build a tree bottom up sampling successively from a bernoulli process. Here are a few examples of such a tree with upto $N = 6$ primitives, *a)* $\cup$ 5 $-$ $\cup$ $\cup$ 2 3 $\cup$ 4 0 1; *b)* $\cup$ $\cup$ 2 3 $-$ $\cup$ 0 1 $\cup$ 4 5; *c)* $\cup$ 0 1; *d)* $\cup$ $\cup$ 1 2 0; or *e)* $\cup$ $\cup$ 2 3 $-$ 0 1.

## 4.4.2 Transformed Primitives

With a count of $N_1$ for single primitives, and $N_2$ for symmetric primitives, we sample the primitives themselves, and their transformation parameters. The symmetric primitives are then cloned, mirrored and inserted in-place to finally generate $N = N_1 + 2N_2$ transformed primitive objects corresponding to the grammar generated in § 4.4.1.

We sample the $N_1 + N_2$ primitives, over a heuristically chosen probability mass function

(PMF) for *{cube, cylinder, prism, sphere, pyramid, cone}*, given as $\{\frac{2}{7}, \frac{2}{7}, \frac{2}{35}, \frac{1}{35}, \frac{1}{35}\}$.

The transformation is split into three parts, translation, scaling and rotation. Rotation is represented as an orientation, *i.e.* one of $\{+X, +Y, +Z, -X, -Y, -Z\}$, sampled over a heuristically chosen PMF given as $\{\frac{1}{16}, \frac{1}{16}, \frac{1}{2}, \frac{1}{16}, \frac{1}{16}, \frac{1}{4}\}$. Scaling and translation are 3-vectors representing one value for each component axis, sampled from a uniform random. The bounds were heuristically fixed as $[0.2, 0.2, 0.2]^T, [1.1, 1.1, 1.1]^T$ for scaling and $[-0.7, -0.7, 0]^T, [0.7, 0.7, 0.3]^T$ for translation.

For each symmetric primitive, the axis of symmetry was sampled from one of $A \sim \{X, Y\}$, and the magnitude of shift was sampled from a gaussian as $r \sim \mathcal{N}(0.4, 0.25)$, so that for translation vector **t** of symmetric primitives, the component along the axis of symmetry be updated as $t_A \leftarrow t_A \pm r$; to yield twice the number of earlier sampled symmetric primitives, $2 \times N_2$.

### 4.4.3 Tonal Art Maps

Akin to cel-shading [KKY01] for discrete color maps, tonal art maps (TAM) were introduced earlier in the area of non-photorealistic animation and rendering (NPAR) to compute the approximate hatch patterns in a sketch [PHWF01]. The core idea may be summarised as follows: *a)* hatch patterns is a line composition, with homogeneous colour and thickness for all lines throughout; *b)* a hatch pattern is stored as a seamlessly tile-able texture, generated procedurally; *c)* for every illumination level a different hatch pattern is generated so that the spatial density of hatch lines approximates the illumination level in grey-scale; *d)* for the purpose of visual continuity of lines, a texture is mapped on to all fragments with illumination lower than the illumination level of the texture itself.

Formally put, let the illumination levels be represented as breakpoints $B \equiv \{b_1, \dots, b_k\}$, and the hatch patterns as texture functions $T \equiv \{\tau_1, \dots, \tau_k\}$, so that $\tau_i(\mathbf{f})$ represents the value of $i$-th texture at fragment **f**. Given a fragment **f** and its phong illumination value as $P(\mathbf{f})$, the TAM shader may be defined as,

$$\text{TAM}(\mathbf{f}; B, T, P) = \prod_{i:b_i \leqslant P(\mathbf{f})} \tau_i(\mathbf{f}) \tag{4.1}$$

Unlike the original TAM, we relaxed on the constraint of seamless tile-ability of the textures, and used hand-drawn hatch patterns as textures.

## 4.5   Model

We leverage the CGAN [GPM+14; MO14] framework. If $F_\theta$ parameterised by $\theta$, be a family of functions modelled as a deep network; $\mathbf{c} \in C$ be a line drawing, $\mathbf{l} \in L$ be the illumination analogy, $\mathbf{t} \in T$ be the set of tonal art maps (aka. textures), and $\mathbf{s} \in S$ be the completed sketch, then the reconstruction loss is defined as in (4.3). For diversity in generation task, $F_\theta$ is implemented with dropouts [KSH12] as implicit noise. In order to bring closer the model-generated sketches and real data distribution, we define a discriminator $D_\phi$ parameterised by $\phi$, as another family of functions modelled by a deep network. The adversarial loss, is thus defined as in (4.4). We hope the model to converge, alternating the optimisation steps for the minimax in (4.2). We call this formulation as **direct model**, illustrated in Fig. 4.6.

In the same spirit, since data generation is cheap (see § 4.4), we see an opportunity to make use of the illumination masks $\mathbf{m} \in M$ as an intermediate step for supervision using a **split model**, for which the losses $\mathscr{L}_{L_1}$ and $\mathscr{L}_{adv}$ in (4.2) are formulated as in (4.5, 4.6).

$$\theta^* = \arg\min_\theta \max_\phi \mathbb{E}_{\mathbf{c},\mathbf{l},\mathbf{t},\mathbf{s}\sim\text{data}} \tag{4.2}$$

$$\left[\mathscr{L}_{L_1}(\theta) + \lambda\mathscr{L}_{adv}(\theta,\phi)\right]$$

**Direct Model:**

$$\hat{\mathbf{s}} = F_\theta(\mathbf{c},\mathbf{l},\mathbf{t})$$

$$\mathscr{L}_{L_1}(\theta) = \|\mathbf{s} - \hat{\mathbf{s}}\|_1 \tag{4.3}$$

$$\mathscr{L}_{adv}(\theta,\phi) = \log(D_\phi(\mathbf{s})) + \log(1 - D_\phi(\hat{\mathbf{s}})) \tag{4.4}$$

**Split Model:**

$$\hat{\mathbf{m}} = F_\theta^{(1)}(\mathbf{c},\mathbf{l})$$

$$\hat{\mathbf{s}} = F_\theta^{(2)}(\hat{\mathbf{m}},\mathbf{t})$$

$$\mathscr{L}_{L_1}(\theta) = \|\mathbf{m} - \hat{\mathbf{m}}\|_1 + \|\mathbf{s} - \hat{\mathbf{s}}\|_1 \tag{4.5}$$

$$\mathscr{L}_{adv}^{(1)}(\theta,\phi) = \log(D_\phi^{(1)}(\mathbf{m})) + \log(1 - D_\phi^{(1)}(\hat{\mathbf{m}}))$$

$$\mathscr{L}_{adv}^{(2)}(\theta,\phi) = \log(D_\phi^{(2)}(\mathbf{s})) + \log(1 - D_\phi^{(2)}(\hat{\mathbf{s}}))$$

$$\mathscr{L}_{adv}(\theta,\phi) = \mathscr{L}_{adv}^{(1)}(\theta,\phi) + \mathscr{L}_{adv}^{(2)}(\theta,\phi) \tag{4.6}$$

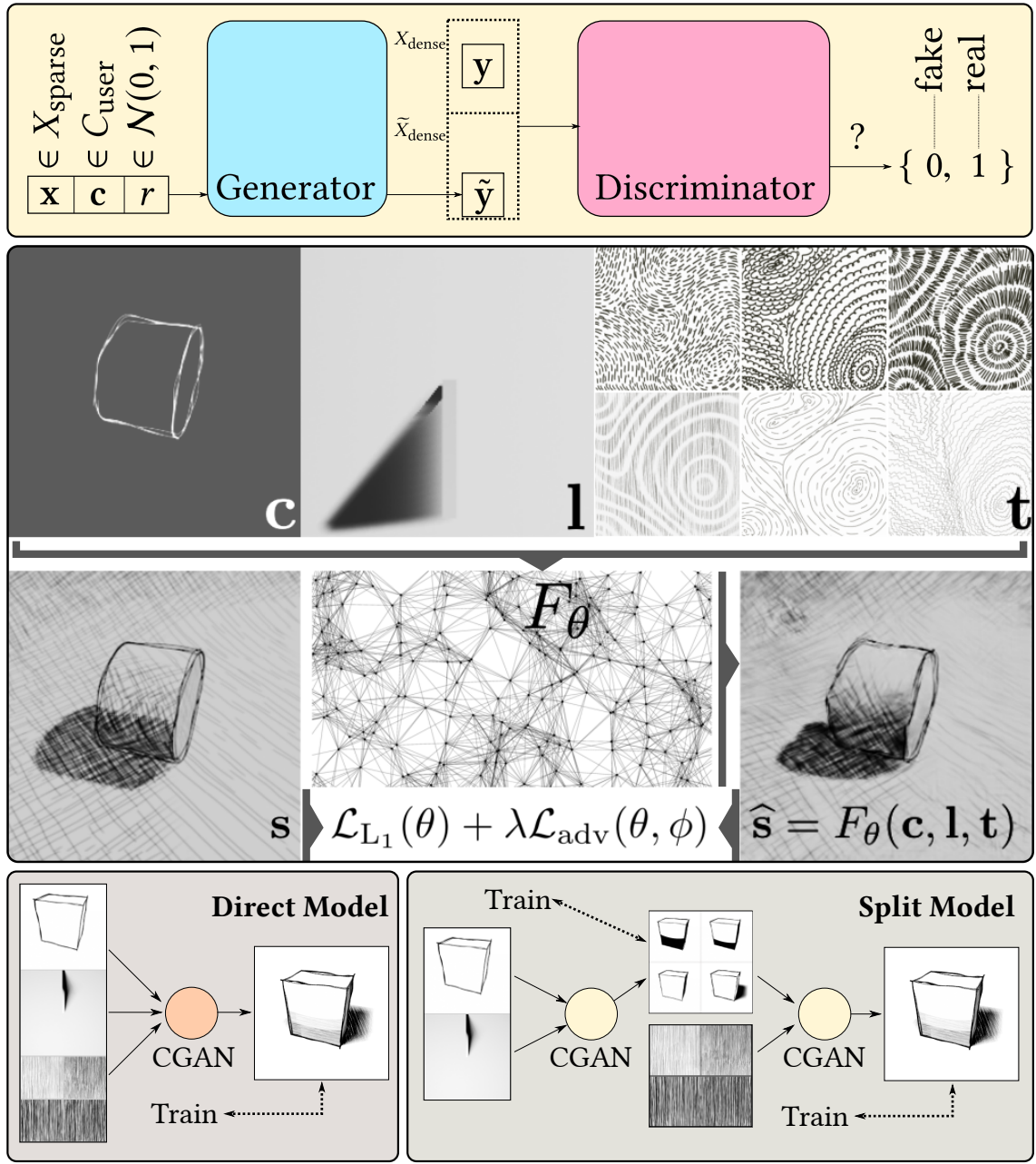

Figure 4.6: The overview of the our framework. *Top to bottom.* Illustration of structure of GAN detailing injection of data from different modalities into the CGAN framework; Further detailed view of direct model; Coarse sketch of a direct model as in (4.3, 4.4, 4.2); Analogous sketch of a split model as in (4.5, 4.6, 4.2).

**The relevance of split model.** The split model features two generators. The former models the conditional probability of cel-shading masks given the contours and the lighting conditions as $P(\mathbf{m}|\mathbf{c},\mathbf{l}) \approx \hat{\mathbf{m}} = F_\theta^{(1)}(\mathbf{c},\mathbf{l})$, and the latter models the conditional probability of the hatch patterns over the input contours, given the cel-shading masks and the artistic textures as $P(\mathbf{s}|\mathbf{m},\mathbf{t}) \approx \hat{\mathbf{s}} = F_\theta^{(2)}(\hat{m},\mathbf{t})$. This gives an opportunity to explicitly supervise the training of masks, are reduces the complexity of the task to be predicting along two modalities, namely the geometry and the coarse effect of lighting, instead of three which would include the detailed appearance too. A similar approach had earlier been seen in StackGAN's [ZXL+17; HLP+17] where the residual was predicted successively so that $\hat{\mathbf{y}} = \sum_i F_1 \circ \cdots \circ F_i(\mathbf{x})$.

## 4.6 Architecture

Akin to earlier attempts like `Pix2Pix` [IZZE16], we utilise U-Net [RFB15] with $\approx$ 12M as our base architecture. Our input comes from three different modalities, *a)* the sketch domain from a human sketching activity; *b)* an illumination hint, *i.e.* a diffuse render of a canonical object under a canonical point of view; and *c)* the texture information as a set of six maps, one corresponding to each tone of brightness/ darkness. To this end, these *multi-modal bounding conditions* are infused into the network by varying the number of input channels, namely the *sparse object outline sketch*, *illumination hint*, and *texture* (See Fig. 4.6). Our hypothesis is that a single network is sufficient to encode the information from the three modalities, that may be further utilised for the downstream task.

We experiment with alternate formulations of the model so that the complete sketch is predicted using a direct model as in (4.3, 4.4), or by predicting an extra intermediate representation of illumination masks using a split model as in (4.5, 4.6). The discriminator is designed as a PatchGAN based classifier, which determines whether $N \times N$ patch in an image is real or fake.

Inspired by recent success of self-attention in adversarial nets [ZLB20], we also use a squeeze-and-excitation [HSS18] based architecture for our model, $F_\theta$ in (4.3, 4.4).

We propose a conditional GAN framework to solve for the task. We adopt the baseline architecture [IZZE16], and further implement a model inspired by a recent advancement as an ablation study. However, our architecture is different and stands out from prior works in the way how muti-modal bounding conditions are being infused into the network. The model so designed can be trained in an end-to-end manner while satisfying the desired constraints.

As a general notion, deep neural networks are difficult to train and are sensitive to the choice of hyper-parameters. But our model doesn't introduce additional complex components, and thus allows us to take advantage of knowing the hyper-parameters of underlying base architecture. This also results in a conceptually clean and expressive model that relies on principled and well-justified training procedures. This is, to the best of our knowledge, the first attempt to solve the problem of generating hatches for sketches while incorporating multiple modalities as constraints.

## 4.7 Experimentation and results

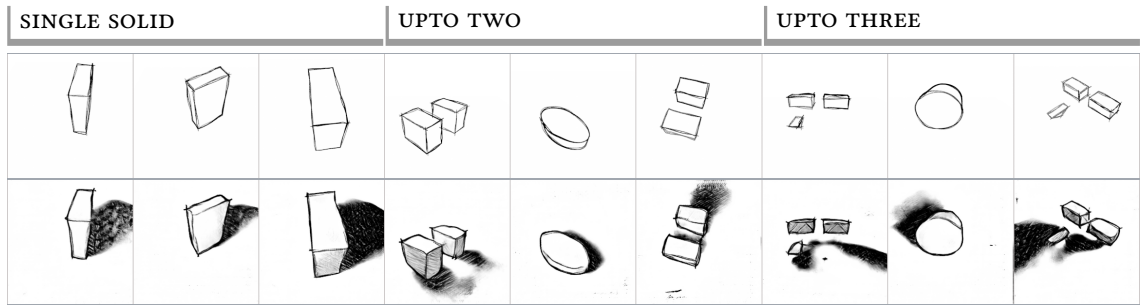

Figure 4.7: Progressive results for DM. Left three with single solid on scene; Middle three with upto two solids; and Rightmost three with upto three solids.

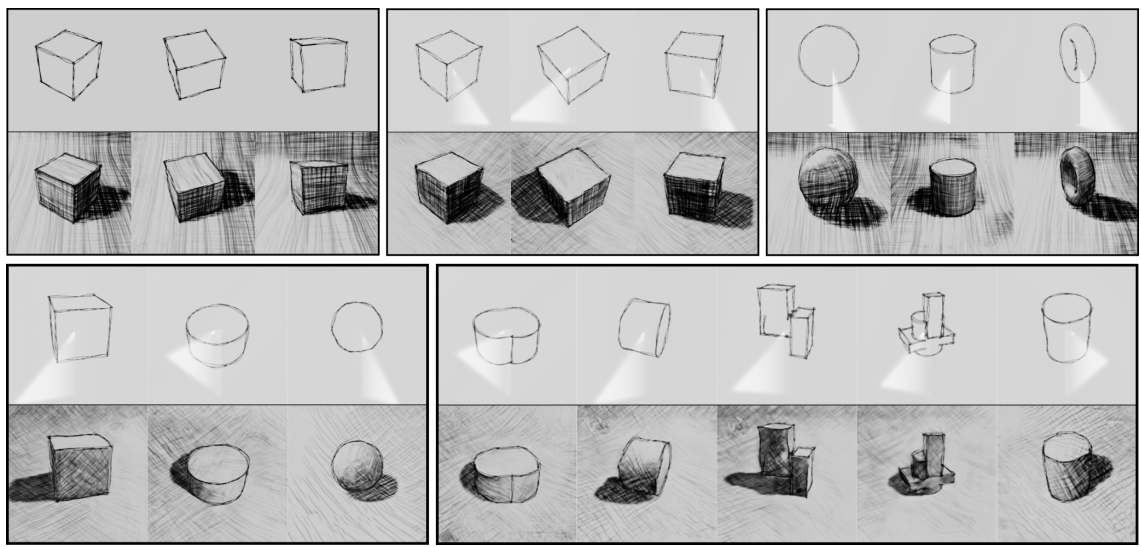

Figure 4.8: Progressive evaluation of DM:WB (see § 4.7) model varying camera pose, illumination, constituent geometry and texture. *Clockwise from top left.* POSE; POSE+LIT; POSE+LIT+SHAP; ALL; TXR. (Details at § 4.7.1.)

IN          DM          SP          SP:WS          SE

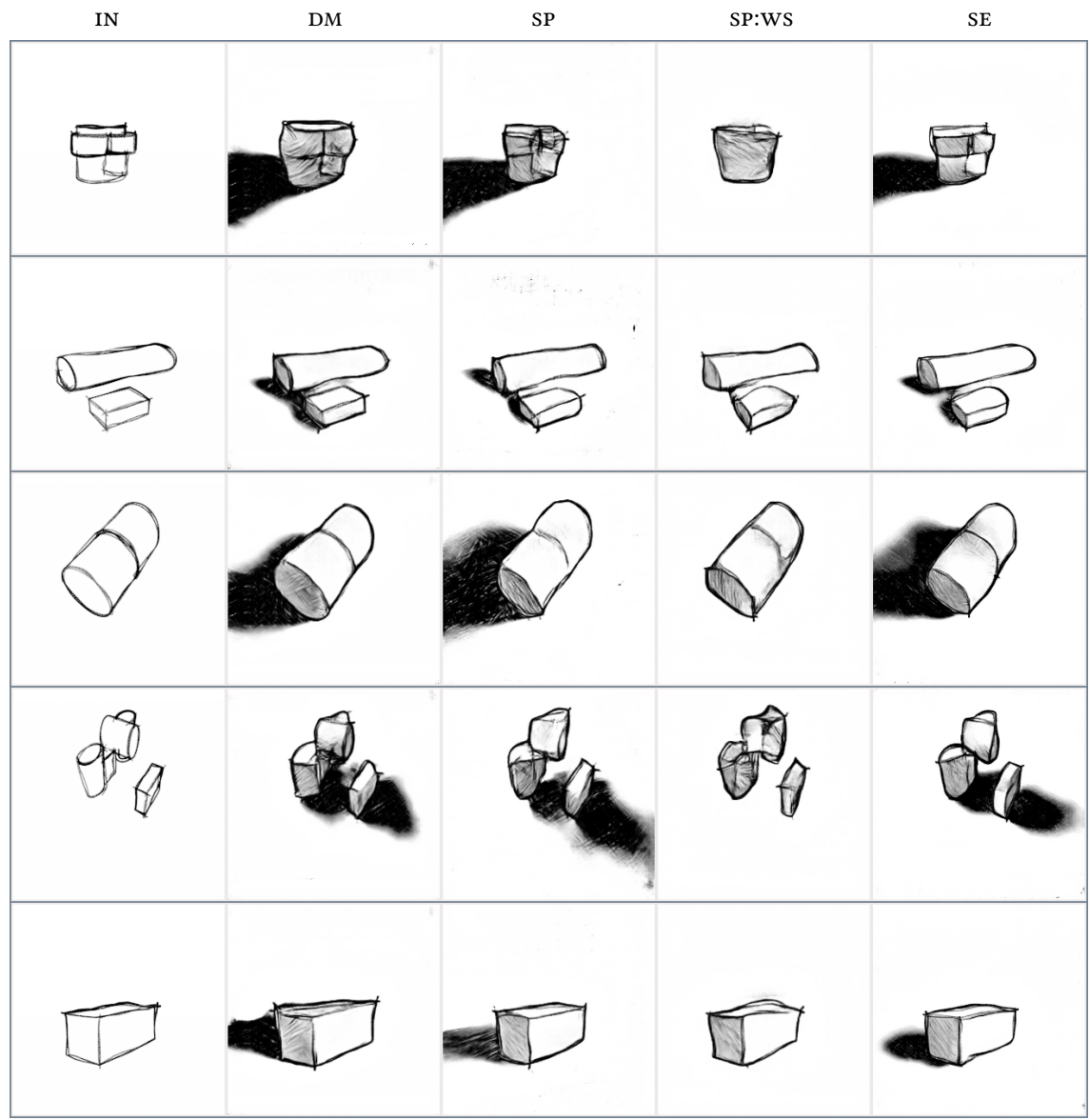

Figure 4.9: Comparative results for all models. *Columns left to right.* Input contour drawing; Corresponding evaluation with DM; With SP; With SP:WS; With SE.

Table 4.1: Quantitative evaluation using PSNR, SSIM, Inference time (in ms), followed by inception scores of model predictions, juxtaposed against those of the dataset.

| MODEL | PSNR | SSIM | Time | Pred. IS | GT IS |
|-------|-------|-------|------|----------|-------|
| DM    | 55.72 | 0.347 | 13   | 4.25     | 5.51  |
| SP    | 55.48 | 0.349 | 33   | 4.12     | 5.51  |
| SE    | 55.90 | 0.376 | 426  | 4.35     | 5.51  |
| SP:WS | 58.92 | 0.287 | 36   | 3.71     | 5.27  |

Table 4.2: Progressive improvement in inception scores against an increase in dataset complexity

| Max objects | 1 | 2 | 3 | 4 | 5 | 6 |
|-------------|------|------|------|------|------|------|
| GT IS       | 3.36 | 3.94 | 4.13 | 4.86 | 5.15 | 5.51 |

To test our hypothesis we performed experiments on the three models, namely *a)* the direct model over U-Net architecture (DM); *b)* the split model over U-Net architecture (SP); *c)* the split model over squeeze-and-excitation architecture (SE). As an extension, we also studied the direct model trained over a dataset with background (DM:WB), and the split model trained over a dataset without shadows on ground (SP:WS). Further, the results are analysed qualitatively (§ 4.7.1), quantitatively (§ 4.7.2) and for generalizability of the model (§ 4.7.3).

### 4.7.1 Qualitative Evaluation

Initially the a DM model was trained with low resolution datasets, starting with single solid in a scene, through to a combination of upto three solids in a scene. The indicative results in Fig. 4.7,show that with progressive increase in the complexity of scene, the model struggles to keep up with the expected response.

To illustrate the strength of our experiments, we present the qualitative results of the following progressive analysis of our DM:WB model with increasingly complex scenes, as shown in Fig. 4.8. The model visibly learns to shade well for given lighting conditions, and standard primitives over a canonical texture. The figure is coded as follows,

POSE: *Pose variations* of a cube, with a canonical scene.

POSE+LIT: *Variations in lighting* and pose of a cube, combined with a canonical texture.

POSE+LIT+SHAP: *Variations in lighting*, pose and shape (single primitive solid in scene), with a canonical texture.

TXR: *Variations in lighting*, pose, shape (single primitive solid in scene) and texture.

ALL: *Variations in lighting*, pose, texture and complex compositions (upto six solids combined in a scene.)

Further qualitative comparison of all the five models, are presented in Fig. 4.9. The nuances of shading with complex solids, and the consequent mistakes, that the model incurs, may escape an untrained eye, solely because, the hatching and shading is acceptable at a coarse level. The models trained without background also perform well but qualitatively seem far from the performance of a model trained with background. Self occlusion and self shadows pose a major challenge.

### 4.7.2 Quantitative analysis

Inception score (IS) [SGZ+16] has been a widely used evaluation metric for generative models. This metric was shown to correlate well with human scoring of the realism of generated images [SGZ+16]. We see in Table 4.2 that as expected the metric increases with increase in scene complexity, indicating progressively more diversity through the experiment.

We evaluate the models, DM, SP, SP:WS, SE, and results are summarised under Table 4.1. In our case we use the NPR rendered images as a reference to the generated sketch completions. The inception scores of the images generated by our models, are comfortably close to the scores for ground truth (GT). Furthermore, the metrics PSNR, SSIM and IS exhibit mild perturbations amongst the models trained with shadow, whereas a sharp change in case of the model trained without shadows. We see a steep decline in the diversity (SSIM and IS), and a sharp ascent in PSNR, both of which can be explained by the absence of shadows, that on one hand limit the scope of expressivity and *diversity*; and on the other, limit the scope of discrepancy from GT and hence *noise*, improving the count of positive signals.

### 4.7.3 Generalizability

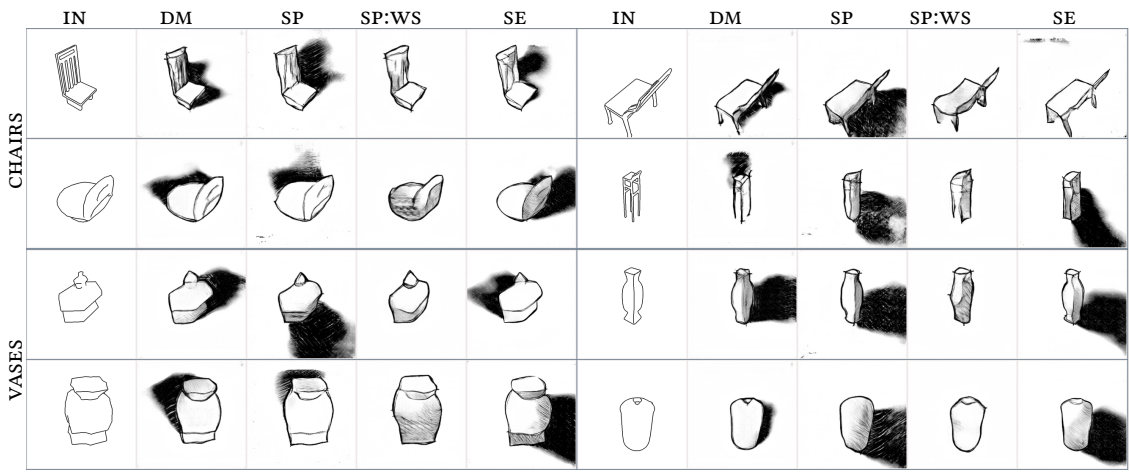

Figure 4.10: Generalizability results — chairs in top two rows; and vases in the next. *Set of five columns left to right.* Input contours; Evaluation with DM; With SP; With SP:WS; With SE.

We deploy two methods to verify a model's tendency to overfit, and/or its ability to generalize. One is to run on other publicly available datasets. To this end, we utilize the class of objects in chairs and vases from the ShapeNet [CFG+15]. And the other method we deploy is by user evaluation (see § 4.8). The representative results of qualitative evauation can be seen in

Fig. 4.10.

### 4.7.4   Limitations

Although the model is able to predict even for human sketches, unseen as well as totally different from the seen dataset, they are barely satisfactory. We have seen here that the model at times modifies the contours, which raises a question of fidelity in production mode. Also, to the best of our knowledge, there seems to be a lack of popularly accepted metric for measuring the levels of realism and diversity in the human sketches. Hence, we borrow the metrics of IS, PSNR, SSIM from optical domain as a proxy to examine and verify a similar trend for our dataset.

## 4.8   User Evaluation

We prepared a tool called DHI—deeply hatch it, for a user to evaluate the model in a real world scenario. The tool leverages the widely popular GIMP program over a pen-touch display to provide the *real-world* digital drawing experience. There is another TK-based applet which allows for choosing the illumination and texture. For consistency, the illumination images were computed in real-time on client-side using the blender backend. The image from the user-end *(client)* is sent to the compute server for inference triggered by a click on the result window on the TK applet, where the response is displayed.

We have classified the interactions as *a)* casual, and *b)* involved. The former was more of a rendezvous with the tool and model. In order to get familiar with the tool, the user was guided by a mentor for a couple of drawings, and later left alone to engage with the tool, hoping for her to develop familiarity and find her comfort zone .

The time of participant engagement varied between 5-43 minutes ($\mu = 21.47$, $\sigma = 5.24$). The sample size was 55 users with their age varying between 12-57 years ($\mu = 26.15$, $\sigma = 7.04$). The results of this evaluation are shown in Fig. 4.11. More impressive results from model trained with background DM:WB have been shown in Fig. 4.1.

We also conducted a more involved group based task completion exercise with 19 volunteers having art/design background, split into groups of 4. This exercise was conducted in three phases and lasted a couple of hours either side of the lunch.

The first phase was an individual level task. A user was required to complete a sketch instead of the computer in a threshold of 40 sec; the threshold had been chosen by consensus. A total of

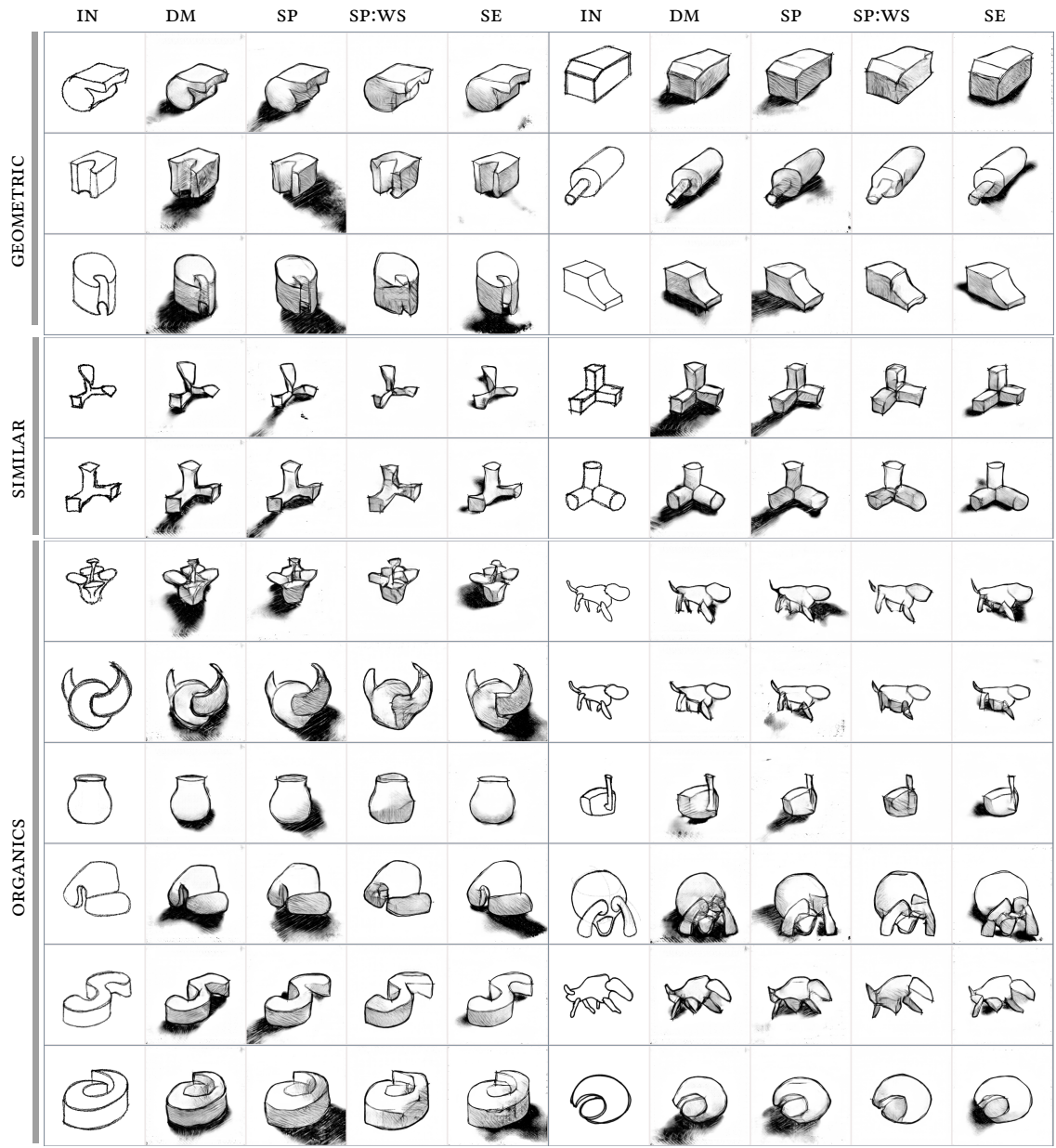

Figure 4.11: User evaluation results. *Blocks top to bottom.* Geometric shapes; Similar shapes with geometric and organic variations; Organic shapes. *Set of five columns left to right.* Input contours; Evaluation with DM; With SP; With SP:WS; With SE.

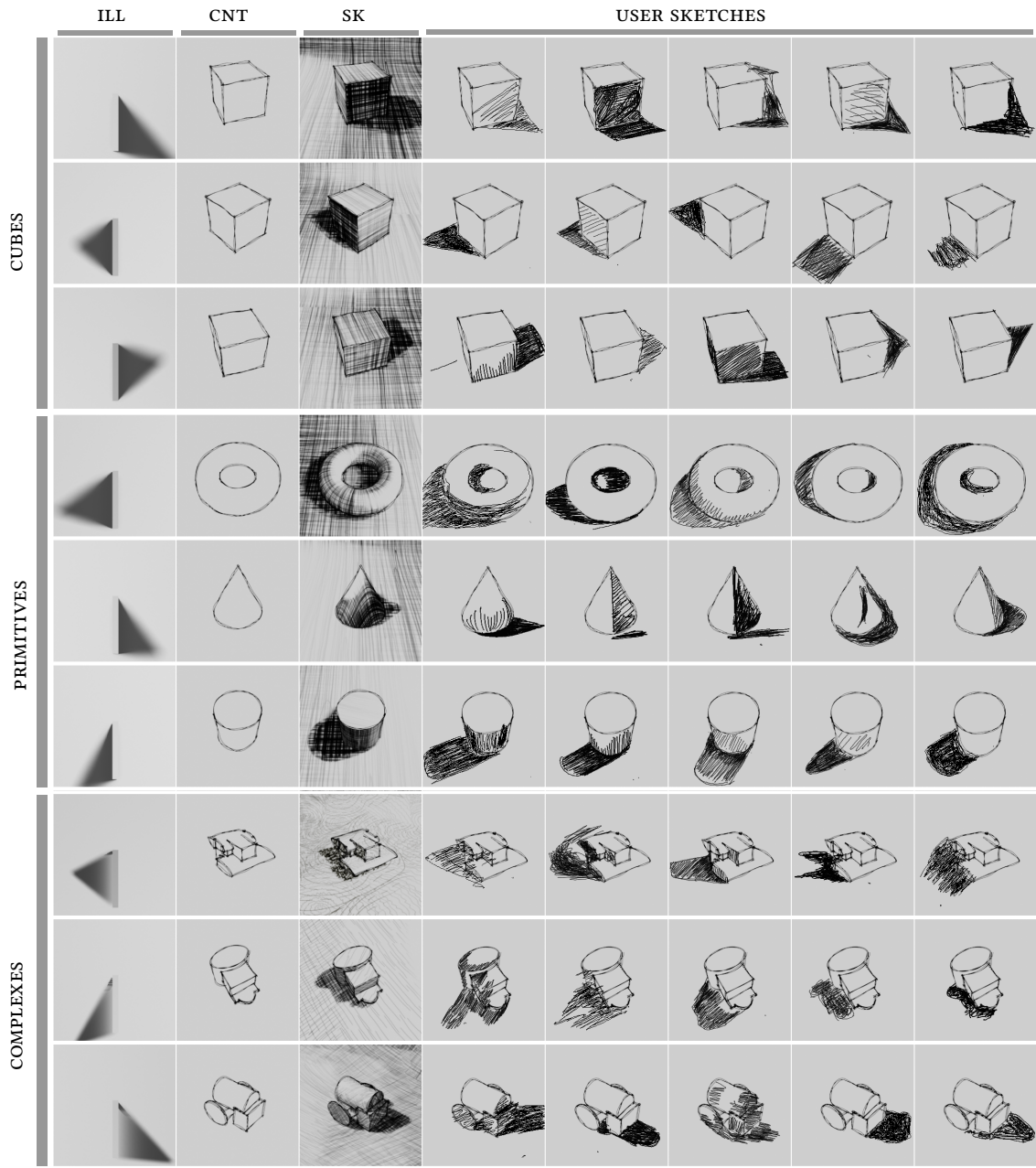

Figure 4.12: Selected responses from user task completion (see § 4.8). *Blocks top to bottom.* With only CUBES in scene; With PRIMITIVES only; With COMPLEX geometries. *Columns left to right.* ILL: Illumination hint; CNT: Contours; SK: GT sketch for reviewers reference; Next five columns show users responses.

21 samples were presented, split into three levels of difficulty: *a)* the first being the nascent level where only a cube is looked at from different poses; *b)* the second consisting a simple geometry like euclidean solids and tori; and *c)* the last were combination of multiple solids generated procedurally. The results shown in Fig. 4.12 indicate that the users seem to have understood the problem principally. The second phase was akin to the casual interaction as described earlier. The last phase, was a group task where a group was required to utilize the tool to create a stop motion picture, one frame at a time.

We encourage the reader to view the video on the project page for the results of this exercise, evaluated against different models here: `https://bvraghav.com/shad3s/`

## 4.9  Discussion

In order to model the function $\mathscr{F} : C \times L \times T \to S$ from § 4.3, we have used the CGAN formulation with a U-Net like architecture compared against an attention framework, thereby showing its effectiveness. Recently more effective architectures, for example using ResNet [HZRS16] blocks instead of sequential convolutions layers, have been presented in the context of image to image translation [CZ20; LCQ+18] by enforcing a cyclic-inference based constraint on the training objective [ZPIE17]. Testing the effectiveness of these architectures and the various training techniques, in the context of GAN-based frameworks for hatch-pattern generation should count as a plausible extension of this work.

The known synthetic factors, namely geometry, illumination conditions, and camera pose; that lead to the data synthesis in § 4.4 also lead to the interesting question of manipulating the latent space, for example observing the latent space arithmetic in DCGAN [RMC15]; and more recently in the context of image to image translation, controlling the latent space in conjunction with the VAE's [KW14] from a Bayesian stand point to synthesise images tailored to the CELEB-A dataset [LKHL20]. The VAE's are vulnerable to mode collapse, and modelling for a domain with sharpness as a preliminary characteristic, like sketches and hatching, is subject to further research. This leaves us with an open problem of investigating the latent space through the recent advances as documented in literature.

Another promising area of recent development is the use of conditional random fields (CRF) over coarse predictions iteratively to converge to a fine level segmentation labels [CPK+18]. The graphical models with energy based formulation over random fields have been shown to be successful with applications involving low frequency details, and also involving sparse signals. Generating hatch patterns, in contrast, is a problem in synthesising a dense high frequency spatial signals, which is an area, open to active research for both Bayesian modelling and deep-learning based methods.

## 4.10   Conclusion

We have shown the use of CGAN in our pipeline for conditional generation of hatch patterns corresponding to line-art representing 3D shapes, by synthesising a large training set, and have portrayed its generalizability over a wide variety of real world user input.

We started with a basic model called DM and experimented with architectures and models with varying complexity, assuming that model capacity was limiting to get more satisfactory results. However, that wasn't the case as suggested by our investigations and as seen from the results; there is no clear winner, in general. From a utilitarian perspective, we should hence suggest the use of simplest and most efficient model, ie. DM.

This is nascent research area, which opens up potential investigation into higher fidelity hatching patterns. Recent advances from the deep learning community have the potential to be explored to improve the results.

CHAPTER 5

# Summary

Sketching as ubiquitous it is as an activity, it is also a different level of intelligence that may be seen when compared to a computation based intelligent system; for a human might look at a projection of the world around, and deploy multiple intelligent systems like acuity, colour, motion *etc.*, collectively called visual intelligence in order to reduce the set of countless possible interpretations to a countable few. And using the same rules of the visual inference system, may "create" the projection of a figment of imagination, that portray a conscious image of this visual system. Intelligent machines on the other hand are philosophically different — designed and trained to solve one problem, and to do so reliably. This thesis had introduced the author's intuitive understanding of the need for an automated geometric inference of sketches, based on a utilitarian standpoint, which have been buttressed with resonating accounts from multiple points of view, as well as with historical evidence in § 1.

**Vectorisation as a graph structure.** The advances in literature that contributed to this thesis either as prior art or as inspiration had been listed out in § 2. The methods for vectorisation may be broadly classified into three as *a)* syntactic or rule based; *b)* stochastic or probabilistic; and *c)* neural processes or deep-learning based. But history suggests that a reliable segmentation, substantially reduces the problem size at the downstream vectorisation step. Taking inspiration from the same, this thesis had proposed to use segmentation as a preceding step to vectorisation and illustrated superlative results.

Although it is common understanding that a sketch may be represented as a planar graph, resolving a sketch as a planar graph of corner points and edges into a graph structure has not been documented to the best of my belief. Thus, with an identified opportunity, the problem of sketch to graph was formulated as a three step pipeline in § 3, with superlative results at the segmentation step. Graph structure was thereafter inferred using a feedback loop, and an industrial application shown through drawing by a robot.

**Application of a graph structure interpretation.** Besides being utilised by a robot for drawing, as illustrated in § 3.4.4, the results may be utilised by any kind of a robot that can perform in a scale agnostic manner. For example, as illustrated by quad-rotor based robots in § 2.3, to stipple on a wall. Or tools other than a pen in a CNC machine may be used to etch, carve, engrave an artistic impression on a solid object.

**Limitations and future scope of this method.** This method is naïve in its graph interpretation step, and showcases the merit of a strong segmentation as a prior to structural inference. Follow-up with improvements in structural interpretation are plausible and desired, but also pluggable into the pipeline.

This method is a batch processing of a sketch. A natural extension may be thought of in the direction of real-time interpretation, the challenges of which may bear similarity to extension along the temporal dimension, *e.g.* the problem of subgraph correspondences.

**Sketch based modelling and completion.** Inferring 3D geometry from 2D drawings is intractable in the general sense, but literature has shown strong correlation between what is drawn, or photographed in 2D, and what is expected in 3D. Researchers have approached to resolve the ambiguity with interactive methods, analytical methods, deep-learning based methods as described in § 2. The need for hatch-pattern based completion of sketches, based on a chosen illumination condition was identified and a gap had been shown to exist in the literature. In § 4, the problem was formulated and solved with the help of a generative framework, trained over a database, that was contributed by this thesis.

**Application of automatic illumination and sketch completion.** Principally, two kinds of situation seem naturally receptive to and in need of an automated solution to sketch assistance and completion, namely *a)* professional use cases with a race against time, *e.g.* book illustrations or caricatures; and *b)* academic use cases where expertise in sketch may not be readily available or affordable, *e.g.* audio-visual presentations in academia.

**Limitations of SHAD3S and future scope.** 3D geometric inference presented in this thesis merely scratches the surface of a vast problem set. The solution suggested here may be critically evaluated from three standpoints.

1. There is a look-ahead bias of having one-to-one correspondence between input shape and inferred geometry. To find many plausible 3D structures satisfying a given 2D contour shape is an open problem, and a possible extension to the problem.

2. The problem here is formulated as for a static image. Extending the work to accommodate the temporal dimension is a natural trajectory to follow. The comes with a gain of multiple views for the same geometry, but also poses a problem of temporal consistency and lower-order geometric correspondences.

3. The solution here results in a latent encoding pertinent to downstream task. Faithfully inferring an explicit 3D geometry, however remains an open problem in literature.

**Augmenting the feedback loop** Recall the discussion on how we intended to impress upon the ability of machine perception to augment the human feedback loop as in Fig. 1.2. The work of 2D interpretation in this thesis is at the very least helpful in carrying out the mundane task of tracing through raw sketches, which we find in many a tasks as an upstream requirement of a design workflow, *e.g.* in an architect's office, in a mural design, in manufacturing and so forth. But more importantly, we may also target higher order tasks like introducing an human-interpretable efficient machine representation of a sketch for downstream requirements, *e.g.* document matching comparing and retrieval all in the sketch domain. The work on 3D interpretation in this thesis involves latent 3D geometric interpretation, and thus provides a convenient benchmark for training a network for desired task.

In this thesis we have documented two different frameworks that may be utilised to decompose a problem involving sketches → geometry into manageable subparts and shown that the extant frameworks in machine-learning may be leveraged to develop, verify and evolve the solution pertinent to the desired task.